\documentclass[runningheads]{llncs}
\usepackage{graphicx}

\usepackage[table]{xcolor}
\usepackage{tikz}
\usepackage{comment}
\usepackage{amsmath,amssymb} %
\usepackage{color}

\usepackage{subcaption}
\newcommand{\etal}{\textit{et al.}}
\usepackage[accsupp]{axessibility}  %
\usepackage{pifont}%
\usepackage{overpic}
\usepackage{booktabs}

\definecolor{amber}{rgb}{1.0, 0.75, 0.0}
\definecolor{darkspringgreen}{rgb}{0.09, 0.45, 0.27}
\usepackage{tabularx}
\usepackage{array}
\usepackage{xfrac}
\newcolumntype{P}[1]{>{\centering\arraybackslash}p{#1}}

\begin{document}
\pagestyle{headings}
\mainmatter
\def\ECCVSubNumber{1064}  %

\title{SinNeRF: Training Neural Radiance Fields on Complex Scenes from a Single Image} %

\titlerunning{SinNeRF: Training NeRF from Single Image} 
\authorrunning{D. Xu, Y. Jiang, et al.} 
\author{%
  Dejia Xu\textsuperscript{1*}, Yifan Jiang\textsuperscript{1*}, Peihao Wang\textsuperscript{1}, Zhiwen Fan\textsuperscript{1}\\ Humphrey Shi\textsuperscript{2,3,4}, Zhangyang Wang\textsuperscript{1}\\
}
\institute{\textsuperscript{1}The University of Texas at Austin, \textsuperscript{2}UIUC, \\\textsuperscript{3}University of Oregon, \textsuperscript{4}Picsart AI Research}

\maketitle

\begin{figure}[h]
    \centering
    \includegraphics[width=0.99\textwidth]{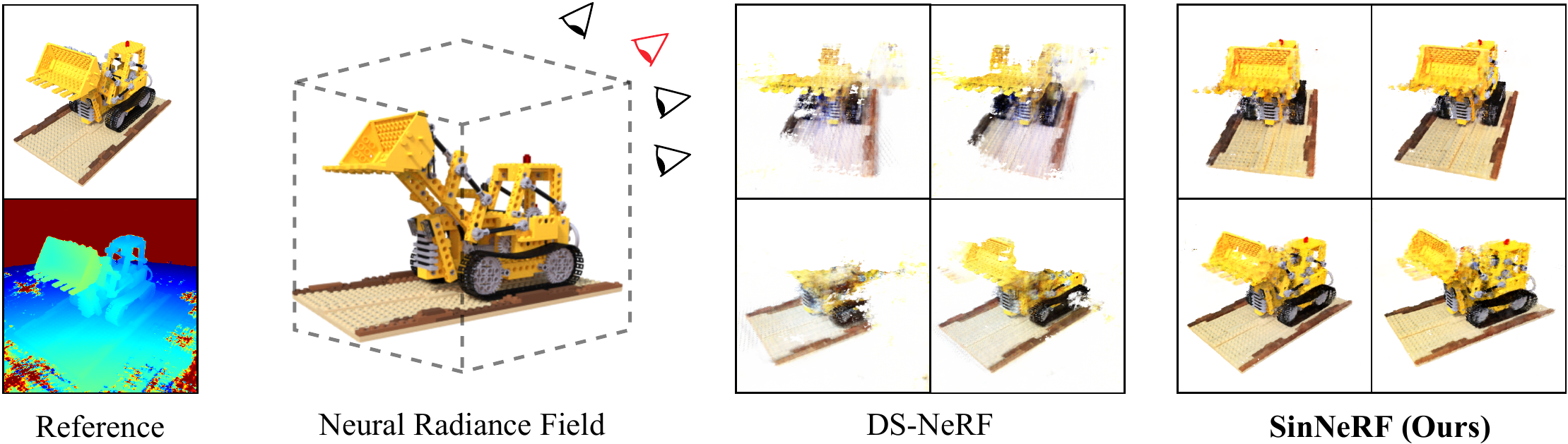}    
\caption{Given only a single reference view as input, our novel semi-supervised framework effectively trains a neural radiance field. In contrast, previous method~\cite{deng2021depth} shows inconsistent geometry when synthesizing novel views.}
\label{fig:teaser}
\end{figure}

\begin{abstract}
Despite the rapid development of Neural Radiance Field (NeRF), the necessity of dense covers largely prohibits its wider applications. While several recent works have attempted to address this issue, they either operate with sparse views (yet still, a few of them) or on simple objects/scenes. In this work, we consider a more ambitious task: training neural radiance field, over realistically complex visual scenes, by ``looking only once", i.e., using only a \textbf{single} view. To attain this goal, we present a \textit{Single View NeRF} (\textbf{SinNeRF}) framework consisting of thoughtfully designed semantic and geometry regularizations. Specifically, SinNeRF constructs a semi-supervised learning process, where we introduce and propagate geometry pseudo labels and semantic pseudo labels to guide the progressive training process. 
Extensive experiments are conducted on complex scene benchmarks, including NeRF synthetic dataset, Local Light Field Fusion dataset, and DTU dataset. We show that even without pre-training on multi-view datasets, SinNeRF can yield photo-realistic novel-view synthesis results. Under the single image setting, SinNeRF significantly outperforms the current state-of-the-art NeRF baselines in all cases. Project page: \url{https://vita-group.github.io/SinNeRF/}
\let\thefootnote\relax\footnote{$\star$ Equal contribution}

\end{abstract}

\section{Introduction}

Synthesizing photo-realistic images has been one of the most essential goals in the area of computer vision. Recently, the field of novel view synthesis has gained tremendous popularity with the success of coordinate-based neural networks. Neural radiance field (NeRF)~\cite{nerf}, as an effective scene representation, has prevailed among image-based rendering approaches. 

Despite its great success, NeRF is impeded by the stringent requirement of the dense views captured from different angles and the corresponding camera poses. As has been implied by recent literature~\cite{niemeyer2021regnerf}., training a neural radiance field without sufficient views will end up with drastic performance degradation, including incorrect geometry and blurry appearance.
Meanwhile, it could be challenging or even infeasible in real-world scenarios to collect a sufficiently dense coverage of views for specific applications such as AR/VR or autonomous driving. Motivated by this, many researchers attempt to address this fragility in the sparse view setting~\cite{yu2021pixelnerf,niemeyer2021regnerf,jain2021putting,kim2021infonerf,deng2021depth,chen2021mvsnerf}. 
One line of research~\cite{yu2021pixelnerf,chen2021mvsnerf} aggregates available learning priors from adequate pre-training on large-scale datasets. Other approaches propose various regularizations on color and geometry of different views~\cite{deng2021depth,jain2021putting,kim2021infonerf,niemeyer2021regnerf}. 
However, most aforementioned works still necessitate multiple view inputs, with a minimum requirement of three views~\cite{niemeyer2021regnerf,deng2021depth}.

In contrast to previous works, we push the setting of sparse views to the extreme by training a neural radiance field on only one \underline{single} view. To our best knowledge, few efforts have been made to explore this circumstance before. PixelNeRF~\cite{yu2021pixelnerf} takes the first attempt by pre-training a feature extractor on a large-scale dataset. Although they report impressive results on simple objects (e.g., ShapeNet dataset~\cite{chang2015shapenet}), their performance on complex scenes~\cite{jensen2014large} is less than satisfactory. Others~\cite{li2021mine,shih20203d} demonstrate good performance on novel-view synthesis. However, their platforms are based on other techniques (e.g., multiplane images). Different from those previous research, our work aims at training the neural radiance field from scratch, without bells and whistles, to generate photo-realistic novel views of complex scenes.

Nevertheless, training a neural radiance field with a single image is frustratingly challenging. \underline{First and foremost}, reconstructing an accurate 3D shape from a single image meets several hurdles. Previous research has addressed reconstructing different types of objects from a single image~\cite{wang2018pixel2mesh,saito2019pifu}.
Especially, Pixel2Mesh~\cite{wang2018pixel2mesh} proposes to reconstruct
the 3D shape from a single image and expressed it in a triangular mesh. PIFu~\cite{saito2019pifu} adopts a 3D occupancy field to recover high-resolution surfaces of humans. NID \cite{wang2022inrdict} utilizes a pre-trained dictionary to acquire implicit fields from sparse measurements. However, all these approaches count on the prior knowledge specific to a certain object class or instance. Thus it can not work for complex scene reconstruction.
\underline{Moreover}, even in the simpler 2D cases, the exploration of training on single images is still gaining much interest as an open problem up to now \cite{shaham2019singan,shocher2018ingan,sitzmann2020implicit,ulyanov2018deep}. 
SIREN~\cite{sitzmann2020implicit} introduces a periodic activation for implicit functions to better fit a single image. SinGAN~\cite{shaham2019singan} and InGAN~\cite{shocher2018ingan} propose to train generative adversarial networks (GANs) using a single image as a reference. Their models can generate visually-pleasing results of images with similar content, but their results often boiled down to approximately replicating or re-composing the patches or textural patterns from the given images, and hence cannot serve the purpose of modeling sophisticated 3D view transformations.

Our inspiration draws from generating pseudo labels according to the available single view, which enables us to design a semi-supervised training strategy to constrain the learned radiance field. Specifically, we design two categories of pseudo labels to capture complementary hidden information. The first one focuses on the geometry of the radiance field, where we reproject depth information between reference view and unseen views through image warping~\cite{huang2021m3vsnet}, thus ensuring multi-view geometry consistency of our trained radiance field. The second one focuses on the semantic fidelity of the unseen views. We utilize a discriminator and a pre-trained Vision Transformer (ViT~\cite{dosovitskiy2020image}) to constrain the unseen views: the former helps improve each unseen view's local textures, while the latter focuses on the perceptual quality of their global structures.

Our main contributions can be summarized as follows:
\begin{itemize}
    \item We propose SinNeRF, a novel semi-supervised framework to train a neural radiance field in complex scenes effectively, using a single reference view.
    \item We introduce and propagate geometry and semantic pseudo labels to jointly guide the progressive training process. The former is inspired by image warping to ensure multi-view geometry consistency, and the latter enforces the perceptual quality of local textures as well as global structures. 
    \item We conduct extensive experiments on complex scene benchmarks and show that SinNeRF can yield photo-realistic novel-view synthesis results without bells and whistles. Under the single image setting, SinNeRF significantly outperforms state-of-the-art NeRF baselines in all cases. 
\end{itemize}

\section{Related Works}

\subsection{Neural Radiance Field}

Neural Radiance Fields (NeRFs)~\cite{nerf} have demonstrated encouraging progress for view synthesis by learning an implicit neural scene representation. Since its origin, tremendous efforts have been made to improve its quality~\cite{verbin2021ref,barron2021mip,barron2021mip360,guo2022nerfren,suhail2022light,chen2022aug}, speed~\cite{muller2022instant,reiser2021kilonerf,sun2022direct,fridovich2022plenoxels}, artistic effects~\cite{wang2022clip,fan2022unified,jain2022zero}, and generalization ability~\cite{chen2021mvsnerf,wang2021ibrnet,liu2022neural,yu2021pixelnerf}. Specifically, Barron \textit{et al.}~\cite{barron2021mip} propose to cast a conical frustum instead of a single ray for the purpose of anti-aliasing. Mip-NeRF 360~\cite{barron2021mip360} further extends it to the unbounded scenes with efficient parameterization. KiloNeRF~\cite{reiser2021kilonerf} speeds up NeRF by adopting thousands of tiny MLPs. MVSNeRF~\cite{chen2021mvsnerf} extracts a 3D cost volume~\cite{yao2018mvsnet,gu2020cascade} and renders high-quality images from novel viewpoints on unseen scenes.
The most related works to SinNeRF target the sparse view setting~\cite{yu2021pixelnerf,deng2021depth,niemeyer2021regnerf,jain2021putting}
Especially, 
DS-NeRF~\cite{deng2021depth} adopts additional depth supervision to improve the reconstruction quality. RegNeRF~\cite{niemeyer2021regnerf} proposes a normalizing flow and depth smoothness regularization. DietNeRF~\cite{jain2021putting} utilizes the CLIP embeddings~\cite{radford2021learning} to add semantic constraints for unseen views.
However, the CLIP embeddings can only be obtained from low-resolution inputs due to memory issues. Thus it struggles to obtain texture details. Meanwhile, these methods can only perform well on at least two or three input views. 
PixelNeRF~\cite{yu2021pixelnerf} utilizes a ConvNets encoder to extract context information by large-scale pre-training, and successfully renders novel views from a single input. However, it can only work on simple objects (e.g., ShapeNet~\cite{chang2015shapenet}) while the results on complex scenes remain unknown.
In our work, we focus on the challenging setting of using only one single view without any pre-training on multi-view datasets.

\subsection{Single View 3D Reconstruction}

Single view 3D reconstruction is a long-standing problem. Early methods use shape-from-shading~\cite{durou2008numerical} or adopt texture~\cite{li2018megadepth} and defocus~\cite{favaro2005geometric} cues. These techniques rely on the existing regions of the images using a depth cue. More recent approaches hallucinate the invisible parts using learned priors. Johnston \textit{et al.}~\cite{johnston2017scaling} adopt an inverse discrete cosine transform decoder.  Fan \textit{et al.}~\cite{fan2017point} directly regresses the point clouds. Wu \textit{et al.}~\cite{wu2017marrnet} learns a mapping from input images to 2.5D sketches and maps the intermediate representations to the final 3D shapes. However, very few datasets are available for 3D annotation, and most of these methods use ShapeNet~\cite{chang2015shapenet} which contains objects of simple shapes.  There are also attempts to reconstruct the 3D shape of specific objects (e.g. humans). PiFU~\cite{saito2019pifu} utilizes a 3D occupancy field to recover the 3D geometry of clothed humans. DeepHuman~\cite{zheng2019deephuman} adopts an image-guided volume-to-volume translation framework. NormalGAN~\cite{wang2020normalgan} conditions a generative adversarial network on the normal maps of the reference view.

Another line of research focuses on learning a 3D representation for view synthesis. Explicit representations involve volumetric representations~\cite{seitz1999photorealistic,henzler2020learning,sitzmann2019deepvoxels}, layer depth images (LDI)~\cite{tulsiani2018layer,shih20203d}, and multiplane images (MPI)~\cite{mildenhall2019local}.  Implicit representations use coordinate-based networks to train a neural scene representation on one single view. PixelNeRF~\cite{yu2021pixelnerf} takes the first attempt by utilizing a pre-trained feature extractor on large-scale dataset. Their results on complex scenes are less than satisfactory compared to their impressive results on simple objects from ShapeNet~\cite{chang2015shapenet}.  GRF~\cite{trevithick2021grf} proposes a generative radiance field modeling 3D geometries by projecting the features of 2D images to 3D points. MINE~\cite{li2021mine} learns a continuous depth MPI and uses volumetric rendering to synthesize novel views. Our work is fundamentally different from existing works in these ways: 1) we train a neural scene representation from scratch without relying on pre-trained feature extractors or multi-plane images; 2) we conduct experiments on complex 3D environments and yield photo-realistic rendered results.

\begin{figure*}[!t]
    \centering
    \includegraphics[width=\textwidth]{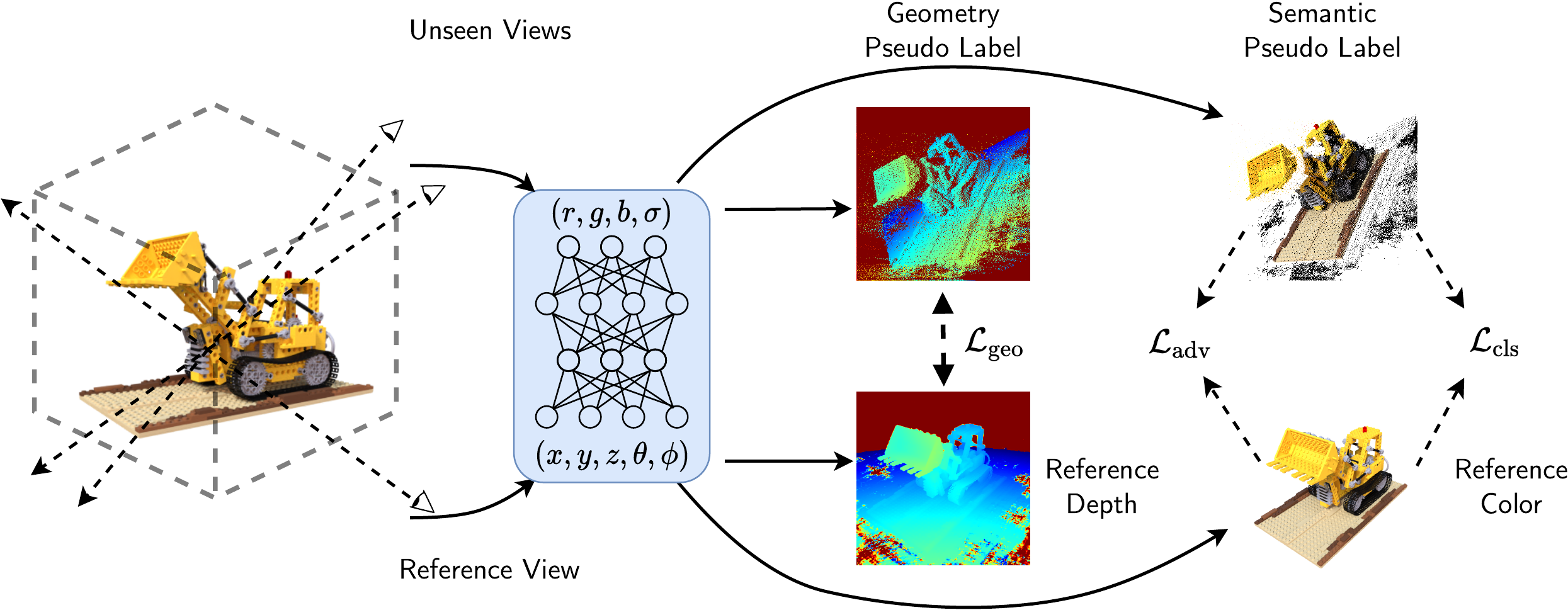}
    \caption{An overview of our SinNeRF, where we synthesize patches from the reference view and unseen views. We train this semi-supervised framework via ground truth color and depth labels of the reference view and pseudo labels on unseen views. We use image warping to obtain geometry pseudo labels and utilize adversarial training as well as a pre-trained ViT for semantic pseudo labels.}
    \label{fig:framework}
\end{figure*}

\subsection{Single Image Training} %

Single image training is a field of great interest in 2D computer vision. SinGAN~\cite{shaham2019singan} and InGAN~\cite{shocher2018ingan} propose a generative adversarial network trained using a single image as reference. Their models can generate visually-pleasing results containing similar content of the image, but the diversity is limited, and their results often copy-paste different patches from the original image. Dmitry \textit{et al.}~\cite{ulyanov2018deep} investigate the deep image prior of convolutional networks and show excellent results in image restoration.  More recently, SIREN~\cite{sitzmann2020implicit} proposes a periodic activation for implicit functions to fit a single image by supervising the gradients of networks. In this work, we make further attempts to adversarially train a radiance field using a single image.

\section{Method}

\subsection{Overview}

The setting of only one single view available is challenging for NeRF, as training directly on the available view leads to overfitting on the reference view and results in a collapsed neural radiance field.
To tackle this problem, we build our SinNeRF as a semi-supervised framework to provide necessary constraints on unseen views. We treat the reference view with RGB and available depth as the labeled set, while the unseen views are considered as the unlabeled set.
To help the neural radiance field render reasonable results on the unseen views, we introduce two types of supervision signals from the perspective of geometry and semantic constraints. We will first introduce the preliminary of neural radiance field and semi-supervised learning framework, then the progressive training strategies.

\subsection{Preliminary}

Neural Radiance Fields (NeRFs)~\cite{nerf} synthesize images sampling 5D coordinates
(location $(x, y, z)$ and viewing direction $(\theta, \phi)$) along camera rays, map them to color $(r, g, b)$ and volume density $\sigma$. Mildenhall \etal~\cite{nerf} first propose to use coordinate-based multi-layer perception networks (MLPs) to parameterize this function and then use volumetric rendering techniques to alpha composite the values at each location and obtain the final rendered images.

Given a pixel $r(t) = o+td$, where $o$ is the camera origin and $d$ is the ray direction, pixel's predicted color is defined as follows:

\begin{equation}
\hat{{C}}({r})=\int_{t_{n}}^{t_{f}} T(t) \sigma({r}(t)) {c}({r}(t), {d}) d t,
\end{equation}
where $T(t)=\exp \left(-\int_{t_{n}}^{t} \sigma({r}(s)) d s\right)$, $\sigma(\cdot)$ and $c(\cdot, \cdot)$ are densities and color predictions from the network. Due to the computational cost, the continuous integral is numerically estimated using quadrature~\cite{nerf}. NeRF~\cite{nerf} optimize the radiance field by minimizing the mean squared error between rendered color and the ground truth color,
\begin{equation}
    \mathcal{L}_{\text{pix}} = \sum_{r \in R_{i}}|| (C(r) - \hat{C}(r)) ||^2,
\end{equation}
where $R_i$ is the set of input rays during training.

\subsection{Geometry Pseudo Label}

Directly overfitting on the reference images leads to a corrupted neural radiance field collapsing towards the provided views. The issue is much more severe when there is only one training image. Without multi-view supervision, NeRF is not able to learn the inherent geometry of the scene and thus fails to build a view-consistent representation. Similar to previous works~\cite{yin2021learning} to reconstruct a 3D shape from a single image, we start by adopting the depth prior to reconstructing reasonable 3D geometry. As suggested by~\cite{deng2021depth}, adding another depth supervision can significantly improve the learned geometry. However, since only a single training view is available in our setting, simply adopting depth supervision can not produce a reasonable 3D shape, as shown in Fig.~\ref{fig:llff}.

To best utilize the available information in the reference view, we propose to propagate it to other views through image warping~\cite{huang2021m3vsnet}. For pixel $p_i(x_i, y_i)$ in reference view $I_\text{ref}$, the corresponding pixel $p_j(x_j, y_j)$ in the $j$-th unseen view $I_\text{unseen}$ can be formulated as:
\begin{equation}
    p_j = K_\text{unseen}T(K^{-1}_\text{ref}Z_ip_i),
\end{equation}
where $Z_i$ is the available depth of reference view, T refers to the relationship between camera extrinsic matrices from $I_\text{ref}$ to $I_\text{unseen}$, and $K_\text{ref}$ and $K_\text{unseen}$ refer to the camera intrinsic matrices.
We further adopt the Painter's Algorithm \cite{newell1972solution}%
when multiple points in the reference view are projected to the same point in the unseen view and select the point with the smallest depth as the warping result.

Through image warping, we then obtain the depth map of an unseen view, which further serves as the pseudo ground truth label. Nevertheless, there is still an unavoidable gap between this pseudo ground truth and its real correspondence, since small misalignment in the predicted depth map can cause large errors when projected to other views. Moreover, it is quite common that the projected results contain some uncertain regions due to the occlusion. To regularize the uncertain regions in the warped results, we utilize the self-supervised inverse depth smoothness loss~\cite{wang2018learning}, which uses the second-order gradients of the RGB pixel value to encourage the smoothness of the predicted depths:

\begin{equation}
\mathcal{L}_{\text {smooth}}\left(d_{i}\right)=e^{-\nabla^{2} \mathcal{I}\left(\mathbf{x}_{i}\right)}\left(\left|\partial_{x x} d_{i}\right|+\left|\partial_{x y} d_{i}\right|+\left|\partial_{y y} d_{i}\right|\right),
\end{equation}
where $d_i$ is the depth map, $\nabla^2 \mathcal{I}(\mathbf{x}_{i})$ refers to the Laplacian of pixel value at location $x_i$.
Similar to ~\cite{wang2018learning}, we calculate this loss on a downscaled resolution.

We also reproject the unseen views back to the reference view to enforce geometry consistency. In summary, the geometry pseudo label is utilized as follows,

\begin{equation}
    \mathcal{L}_{\text{geo}} = \mathcal{L}_1(d_1, f(d_2)) + \mathcal{L}_1(f(d_1), d_2) + \lambda_4 \mathcal{L}_{\text{smooth}},
\end{equation}
where $\lambda_4$ is empirically set to be 0.1 in all our experiments, $d_1$ and $d_2$ refer to the depths of two views, and $f(\cdot)$ refers to the image warping result of the other view using the current view's depth information.

\subsection{Semantic Pseudo Label}

Since the rendered color and texture might still be inconstant across different views, image warping can only project depth information. We propose to adopt semantic pseudo labels to regularize the learned appearance representation.
Unlike the geometry pseudo labels, where we enforce the consistency in 3D space, semantic pseudo labels are adopted to regularize the 2D image fidelity. Concretely speaking, we introduce a local texture guidance loss implemented by adversarial learning, and a global structure prior supported by a pre-trained ViT network. The two complementary guidances collaboratively help SinNeRF render visually-pleasing results in each view.

\subsubsection{Local Texture Guidance}

The local texture guidance is implemented via a patch discriminator. The outputs from NeRF are considered as fake samples, and the patches randomly cropped from the reference view are regarded as real samples. Since the available training data are too limited, the discriminator tends to memorize the entire training set. To overcome this issue, we adopt differentiable augmentation~\cite{zhao2020diffaugment} for our discriminator to improve its data efficiency:
\begin{equation}
\begin{aligned}
\mathcal{L}_\text{D} &=\mathbb{E}_{\boldsymbol{x} \sim p_{\text {data }}(\boldsymbol{x})}\left[f_{D}(-D(T(\boldsymbol{x})))\right] +\mathbb{E}_{\boldsymbol{z} \sim p(\boldsymbol{z})}\left[f_{D}(D(T(G(\boldsymbol{z}))))\right], \\
\mathcal{L}_\text{G} &=\mathbb{E}_{\boldsymbol{z} \sim p(\boldsymbol{z})}\left[f_{G}(-D(T(G(\boldsymbol{z}))))\right], \\
 \mathcal{L}_\text{adv} &= \mathcal{L}_\text{D} + \mathcal{L}_\text{G},
\end{aligned}
\end{equation}
where $T$ refers to the augmentation applied on both real and fake samples. We train the GAN framework using Hinge loss~\cite{lim2017geometric}, so $f_D(x)=\max(0,1+x)$ and $f_G(x)=x$. 
The architecture of our discriminator is a cascade of convolutional layers. More details about the discriminator design is provided in the supplementary materials.

\subsubsection{Global Structure Prior}

Vision transformers (ViT) have been proven to be an expressive semantic prior, even between images with misalignment~\cite{tumanyan2022splicing,amir2021deep}.  Similar to ~\cite{jain2021putting},
we propose to adopt a pre-trained ViT for global structure guidance, which enforces semantic consistency between unseen views and the reference view. Although there exists pixel-wise misalignment between the views, we observe that the extracted representation of ViT is robust to this misalignment and provides supervision at the semantic level.
Intuitively, this is because the content and style of the two views are similar, and a deep network is capable of learning invariant representation.

Here we adopt DINO-ViT~\cite{caron2021emerging}, a self-supervised vision transformer trained on ImageNet~\cite{deng2009imagenet} dataset. Unlike DietNeRF~\cite{jain2021putting} which utilizes a CLIP-ViT~\cite{radford2021learning} and adopts its projected images embeddings as features, we directly extract the [CLS] token from DINO-ViT's output. This approach is more straightforward since the [CLS] token serves as a representation of an entire image~\cite{dosovitskiy2020image}. The intuition also aligns with the recent findings of~\cite{tumanyan2022splicing}, where ViT architecture can capture semantic appearance after self-supervised pre-training. We calculate $L_2$ distance between the extracted features,

\begin{equation}
    \mathcal{L}_{\text{cls}} = ||f_{\text{vit}}(A) - f_{\text{vit}}(B) ||^2
\end{equation}
where $f_{\text{vit}}(\cdot)$ refers to the extracted [CLS] tokens. $A$ and $B$ are patches from the reference view and an unseen view, respectively.

\subsection{Progressive Training Strategy}
To stabilize the training of the GAN framework,
we apply a progressive sampling strategy to the training of a single view neural radiance field.

\paragraph{\textbf{Progressive Strided Ray Sampling:}}
We start from utilizing a stride sampling~\cite{schwarz2020graf} of ray generation and progressively reduce the stride size during training. This design enables our SinNeRF to cover a much larger region with a limited amount of rays. Specifically, the $K \times K$ patch $P$ of stride $s$ containing point $(u, v)$ is defined as a set of 2D image coordinates,

\begin{equation}
\mathcal{P}(u, v, s)=\left\{(u + sx, v +sy) \mid x, y \in\left\{0, \ldots, {K}\right\}\right\}.
\end{equation}
Under this circumstance, the NeRF is able to generate a $K \times K$ patch representing a large aspect of the scene. During training, we randomly sample two patches in each iteration, with the first one from the reference view and the other one from a random unseen view. After that, the collaborative local texture guidance and global structure prior loss are applied on these patches to provide semantic guidance on the unseen views. Meanwhile, we obtain the geometry pseudo labels via image warping and add regularization on each patch's intersection with the corresponding patch's warped result. As the training goes into the latter stages, the stride $s$ decreases so that the framework starts to focus on more local regions. 
Note that we randomly initialize the discriminator after reducing the stride size. This helps the discriminator focus on a fixed resolution, making the training more stable.

\paragraph{\textbf{Progressive Gaussian Pose Sampling:}} After that, we propose to progressively enlarge the viewing angle during training. During training, we start at a local neighbor of the reference view and progressively rotate the camera pose more as the training proceeds. This helps the network to focus on dealing with the confident regions and stabilize training as the output image patches will have a good quality when the camera pose is only slightly different from the reference view. Specifically, we represent the distance between an unseen view and the reference view as Euler angles. Let $(\alpha, \beta, \phi)$ denote the signed angles between the axis in the reference view's camera coordinate and the axis in the unseen views' camera coordinates. In each iteration, we sample $\alpha, \beta, \phi$ each based on a Gaussian distribution $\mathcal{N}(0, \omega^2)$, where $\omega$ increases with more iterations.

We show the overall loss function as follows:
\begin{equation}
    \mathcal{L}_{\text{total}} = \mathcal{L}_{\text{pix}} + \lambda_1 \mathcal{L}_{\text{geo}} + \lambda_2 \mathcal{L}_{\text{adv}} + \lambda_3 \mathcal{L}_{\text{cls}},
\end{equation}
where $\lambda_1,\lambda_2,\lambda_3$ are weighting factors. We anneal the loss weight during training. In the early stages where we use a large stride and the patch covers the major regions of the original image, the global structure prior is given a large weight $\lambda_3$ compared to the weight of local texture guidance $\lambda_2$. As the training proceeds, we reduce the stride to focus on reconstructing the high-frequency details. Consequently, we reduce the weight of global structure prior $\lambda_3$ and increase the weight of local texture guidance $\lambda_2$. In all our experiments, $\lambda_1$, $\lambda_2$, and $\lambda_3$ are initialized to be 8, 0.1, and 0, respectively. During the training process, we gradually decrease $\lambda_2$ to 0 and increase $\lambda_3$ to 0.1 with a linear function.

\section{Experiment}

\subsection{Implementation Details}

We use the same architecture as the original NeRF paper~\cite{nerf}. During training iterations, we randomly sample two patches of rays from both the reference view and a random sampled unseen view. The size of patches on NeRF synthetic (Blender) dataset, Local Light Field Fusion (LLFF) dataset, and DTU dataset are set as $64 \times 64$, $84 \times 63$, and $70 \times 56$, respectively. The rendered patches are then sent to the discriminator and DINO-ViT network, where we additionally resize its input patches to $224\times 224$ resolution to fit the input resolution of DINO-ViT architecture.
We train our framework using RAdam optimizer~\cite{liu2019radam}, with an initial learning rate of $1e-3$. We decay the learning rate by half after every 10k iterations. The learning rate of the discriminator is kept to be $20\%$ of the MLP's learning rate. The stride for sampling the patches starts at 6 and gradually reduces by 2 after every 10k iterations.
All experiments of SinNeRF are conducted on an NVIDIA RTX A6000 GPU. The whole training process takes several hours for each scene. More implementation details and visual results are provided in the supplementary.

\begin{figure*}[!t]
    \centering
    \begin{tabular}{P{0.19\textwidth}P{0.19\textwidth}P{0.19\textwidth}P{0.19\textwidth}P{0.19\textwidth}}
    \scriptsize DS-NeRF~\cite{deng2021depth} & \scriptsize PixelNeRF~\cite{yu2021pixelnerf} & \scriptsize DietNeRF~\cite{jain2021putting} & \scriptsize  SinNeRF & \scriptsize  Target Image\\
    \end{tabular}
    \vskip\baselineskip
    \begin{subfigure}[b]{\textwidth}
        \centering
        \includegraphics[width=0.19\linewidth]{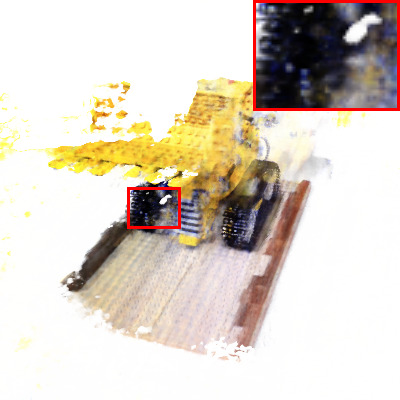}
        \includegraphics[width=0.19\linewidth]{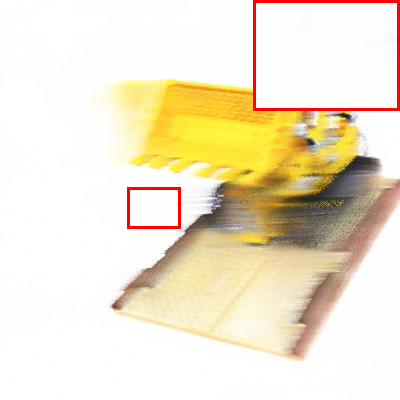}
        \includegraphics[width=0.19\linewidth]{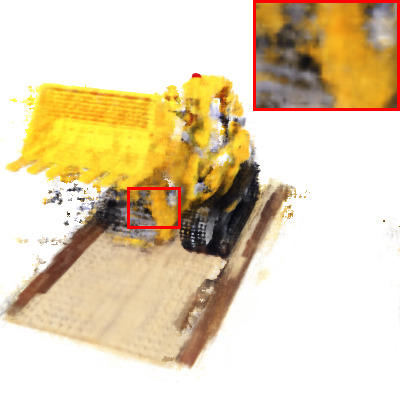}
        \includegraphics[width=0.19\linewidth]{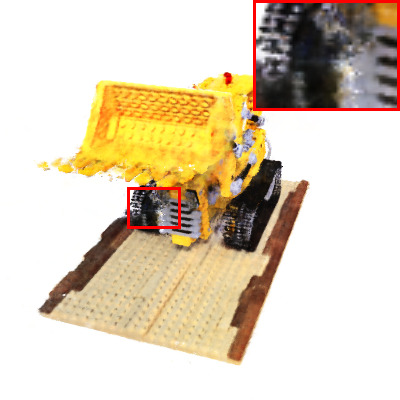}
        \includegraphics[width=0.19\linewidth]{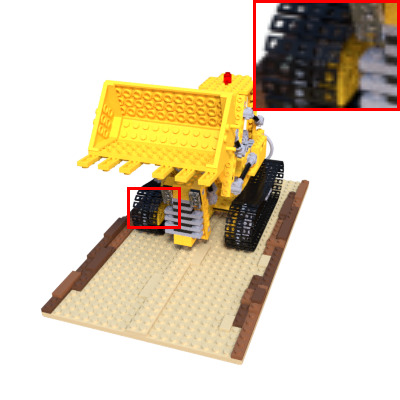}
    \end{subfigure}
    \begin{subfigure}[b]{\textwidth}
        \centering
        \includegraphics[width=0.19\linewidth]{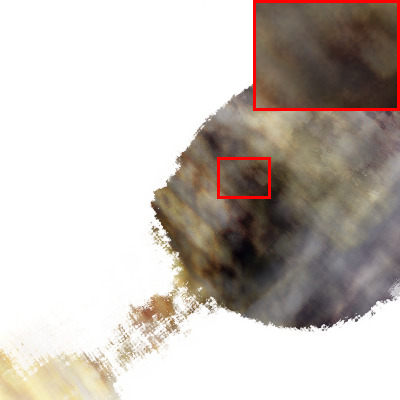}
        \includegraphics[width=0.19\linewidth]{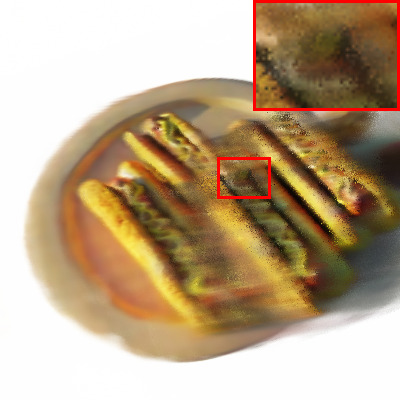}
        \includegraphics[width=0.19\linewidth]{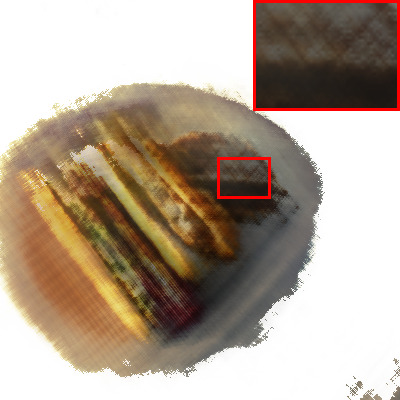}
        \includegraphics[width=0.19\linewidth]{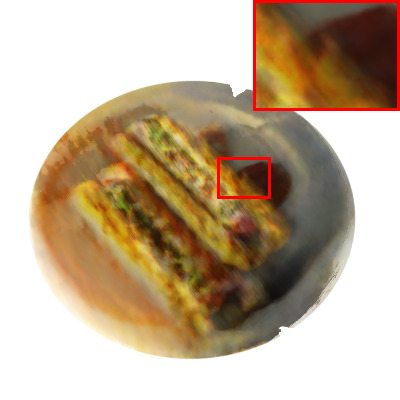}
        \includegraphics[width=0.19\linewidth]{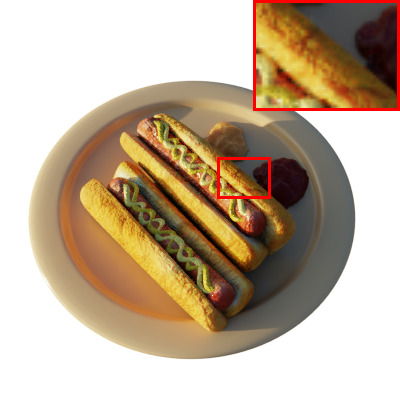}
    \end{subfigure}
    \begin{subfigure}[b]{\textwidth}
        \centering
        \includegraphics[width=0.19\linewidth]{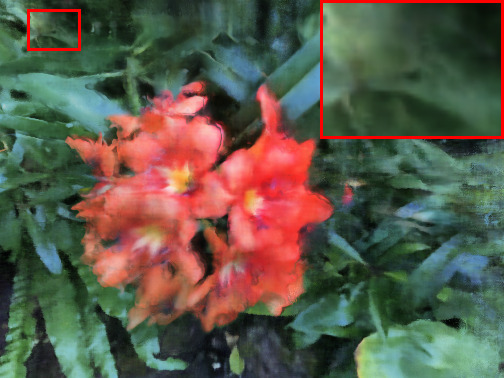}
        \includegraphics[width=0.19\linewidth]{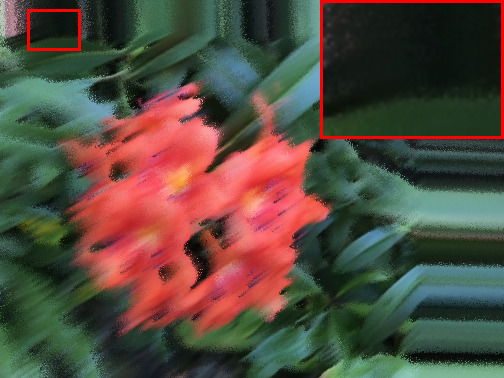}
        \includegraphics[width=0.19\linewidth]{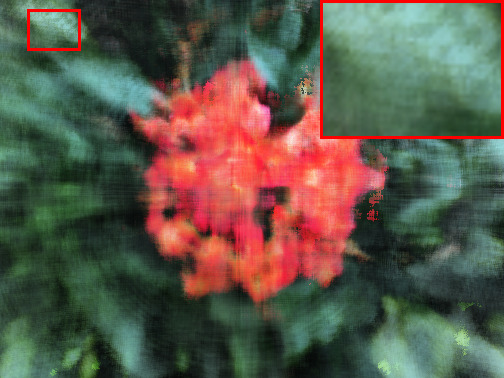}
        \includegraphics[width=0.19\linewidth]{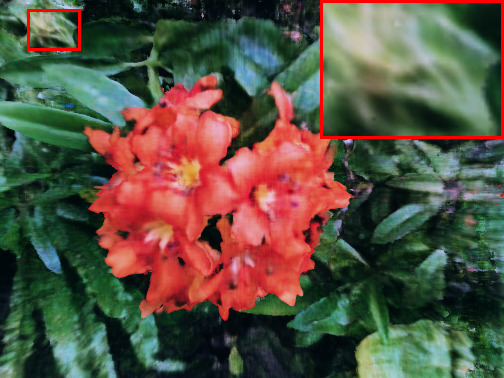}
        \includegraphics[width=0.19\linewidth]{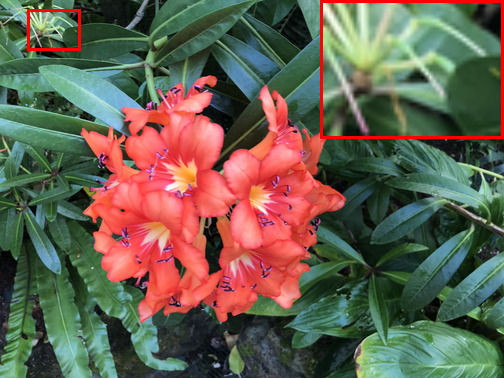}
    \end{subfigure}
        \begin{subfigure}[b]{\textwidth}
        \centering
        \includegraphics[width=0.19\linewidth]{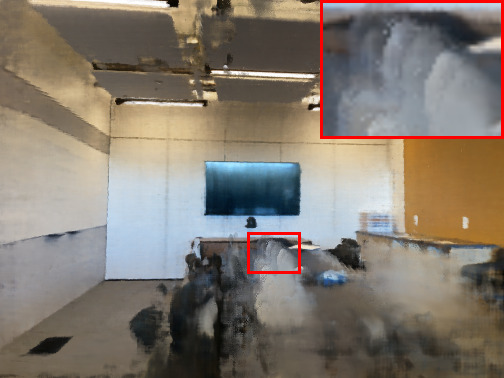}
        \includegraphics[width=0.19\linewidth]{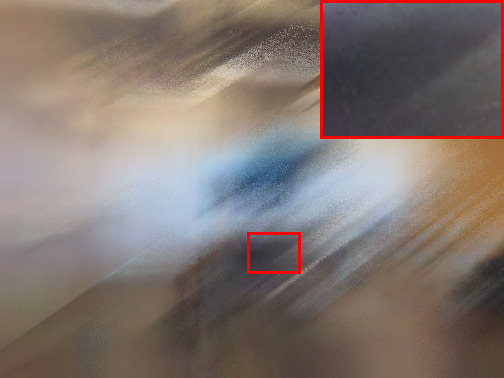}
        \includegraphics[width=0.19\linewidth]{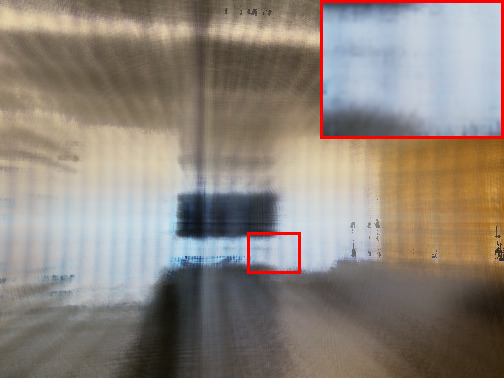}
        \includegraphics[width=0.19\linewidth]{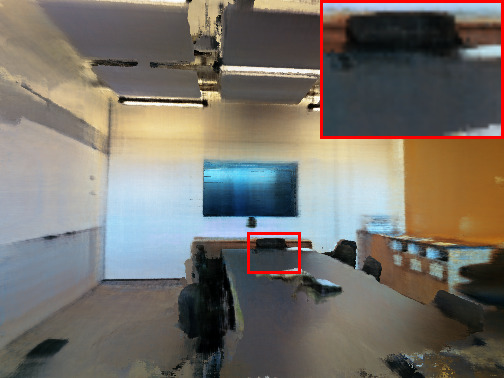}
        \includegraphics[width=0.19\linewidth]{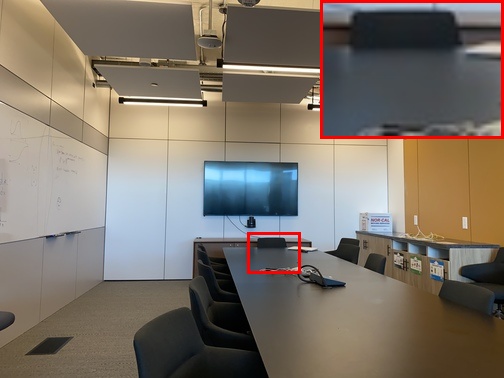}
    \end{subfigure}
\caption{Novel view synthesis results of different methods on NeRF synthetic and LLFF Dataset.}\label{fig:llff}
\end{figure*}

\begin{figure}[!h]
    \centering
    \begin{tabular}{P{0.19\textwidth}P{0.19\textwidth}P{0.19\textwidth}P{0.19\textwidth}P{0.19\textwidth}}
    \scriptsize DS-NeRF~\cite{deng2021depth} & \scriptsize PixelNeRF~\cite{yu2021pixelnerf} & \scriptsize DietNeRF~\cite{jain2021putting} & \scriptsize  SinNeRF & \scriptsize  Target Image\\
    \end{tabular}
    \begin{subfigure}[h]{\textwidth}
        \centering
        \includegraphics[width=0.19\linewidth]{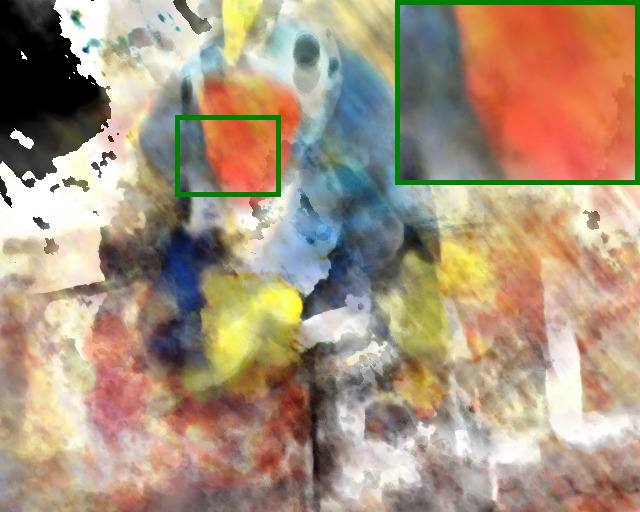}
        \includegraphics[width=0.19\linewidth]{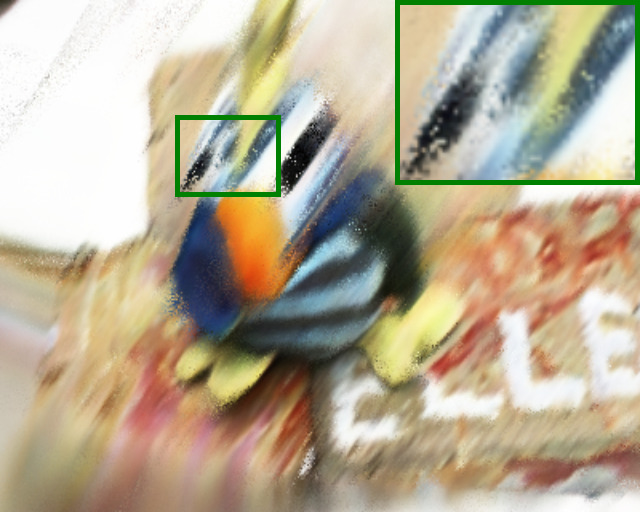}
        \includegraphics[width=0.19\linewidth]{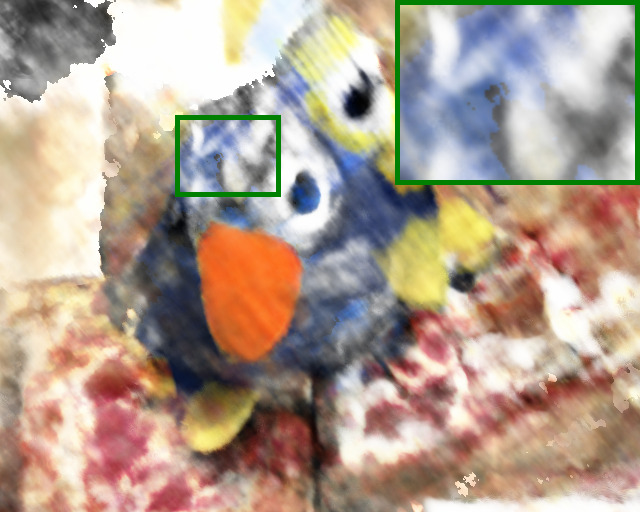}
        \includegraphics[width=0.19\linewidth]{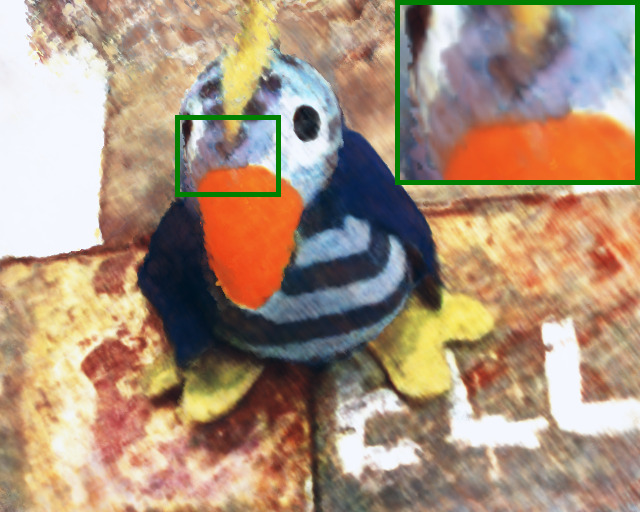}
        \includegraphics[width=0.19\linewidth]{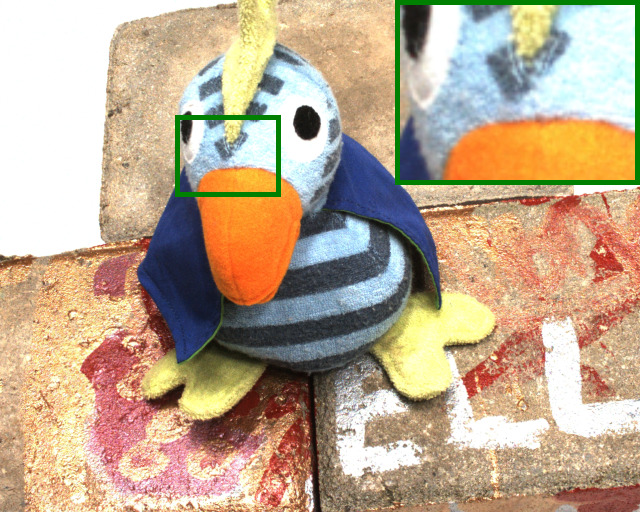}
    \end{subfigure}
        \begin{subfigure}[h]{\textwidth}
        \centering
        \includegraphics[width=0.19\linewidth]{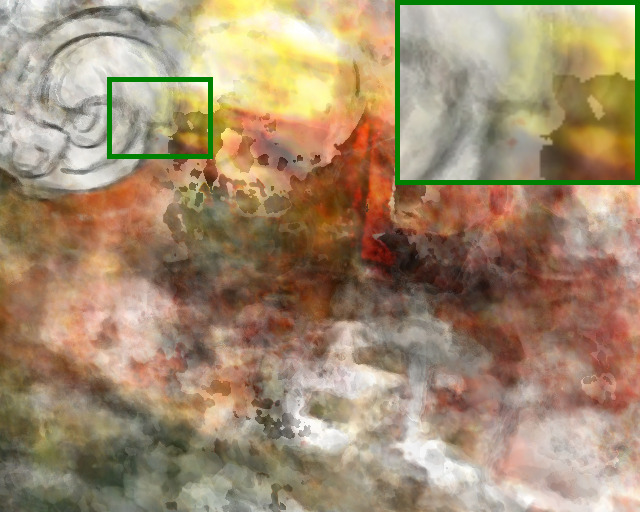}
        \includegraphics[width=0.19\linewidth]{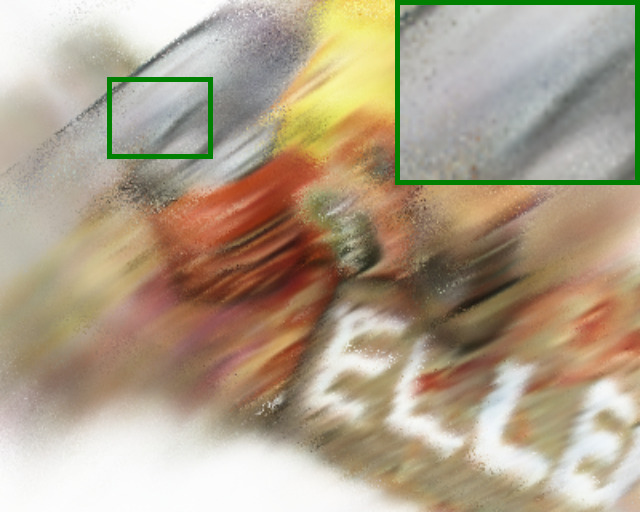}
        \includegraphics[width=0.19\linewidth]{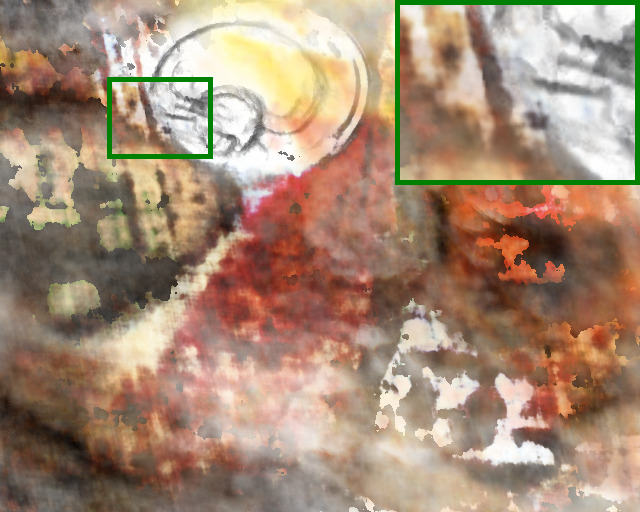}
        \includegraphics[width=0.19\linewidth]{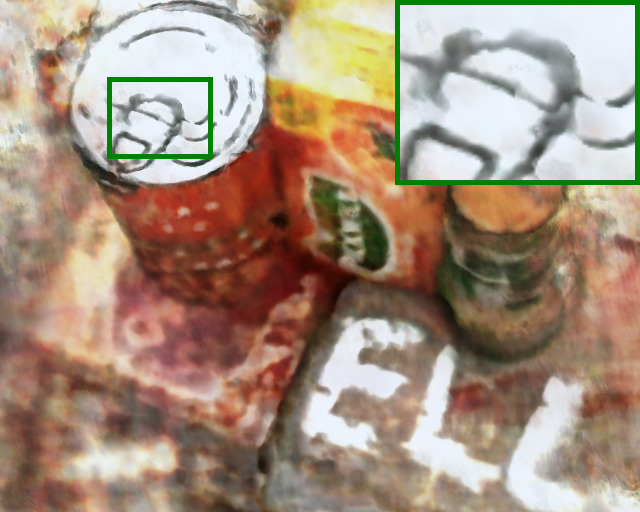}
        \includegraphics[width=0.19\linewidth]{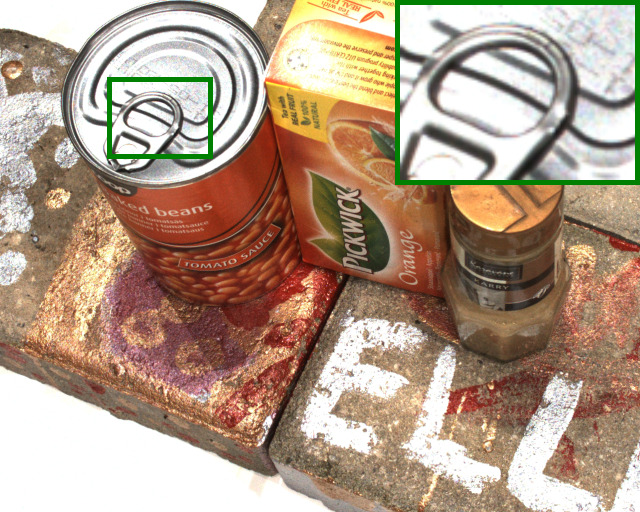}
    \end{subfigure}
    \begin{subfigure}[h]{\textwidth}
        \centering
        \includegraphics[width=0.19\linewidth]{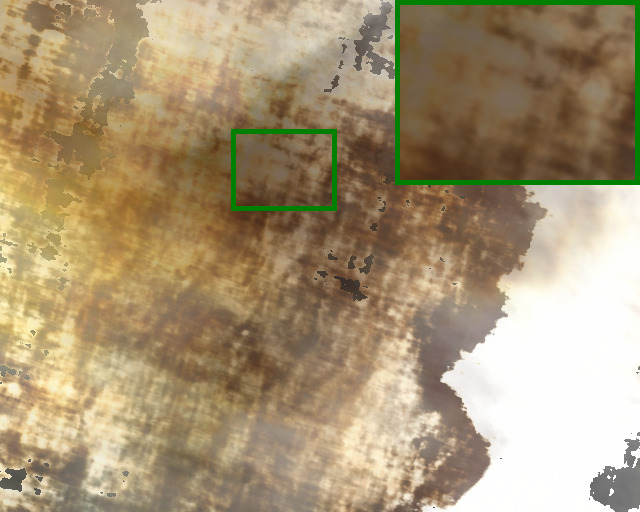}
        \includegraphics[width=0.19\linewidth]{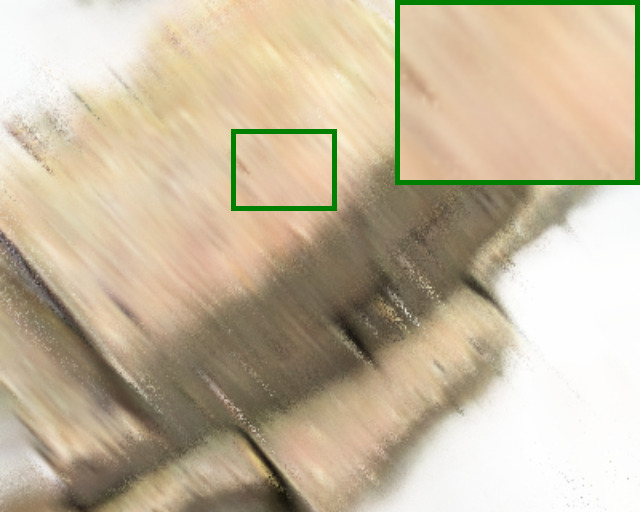}
        \includegraphics[width=0.19\linewidth]{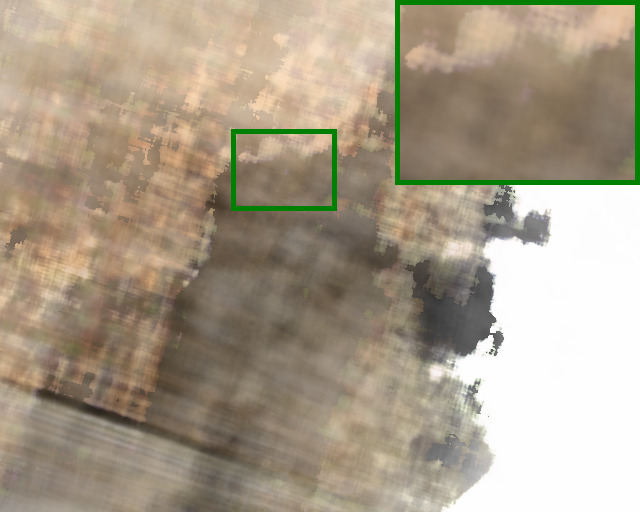}
        \includegraphics[width=0.19\linewidth]{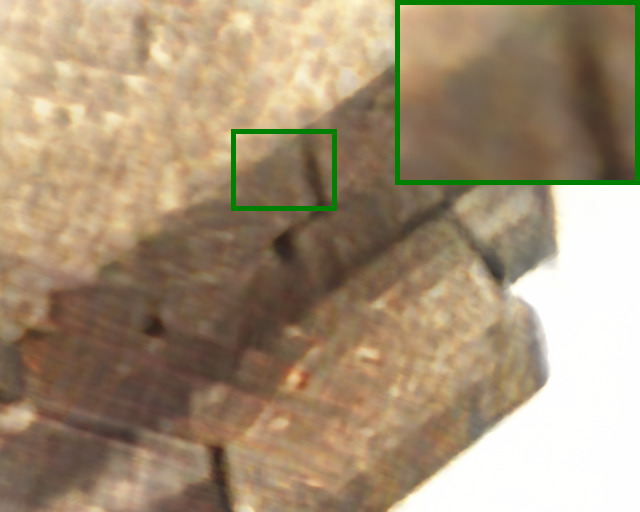}
        \includegraphics[width=0.19\linewidth]{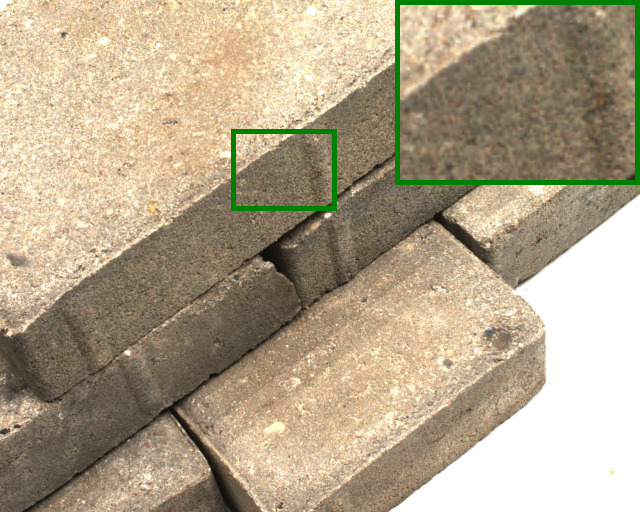}
    \end{subfigure}
    \begin{subfigure}[h]{\textwidth}
        \centering
        \includegraphics[width=0.19\linewidth]{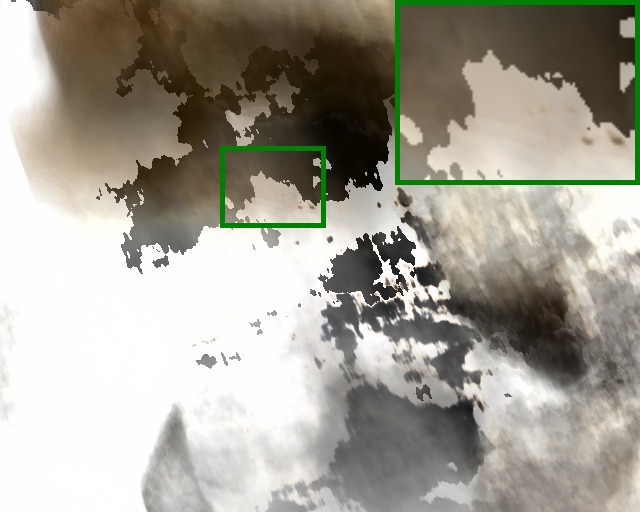}
        \includegraphics[width=0.19\linewidth]{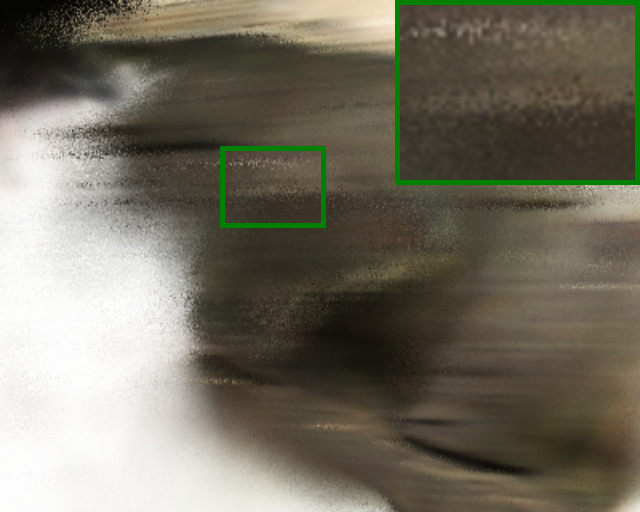}
        \includegraphics[width=0.19\linewidth]{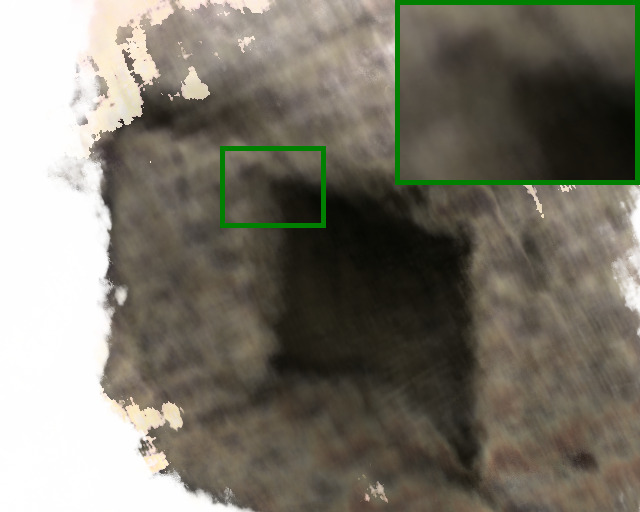}
        \includegraphics[width=0.19\linewidth]{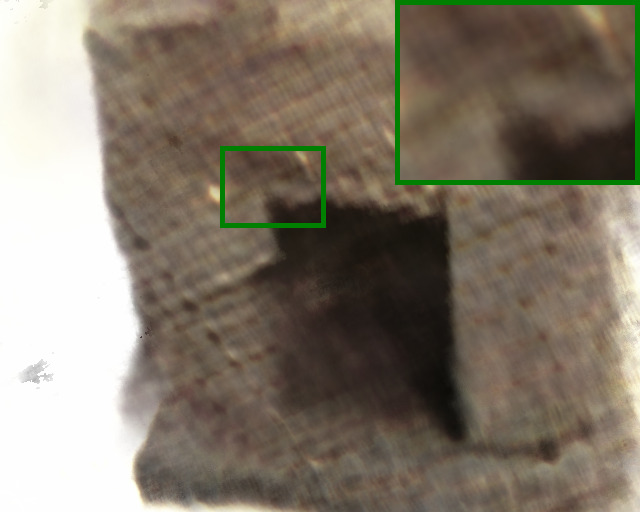}
        \includegraphics[width=0.19\linewidth]{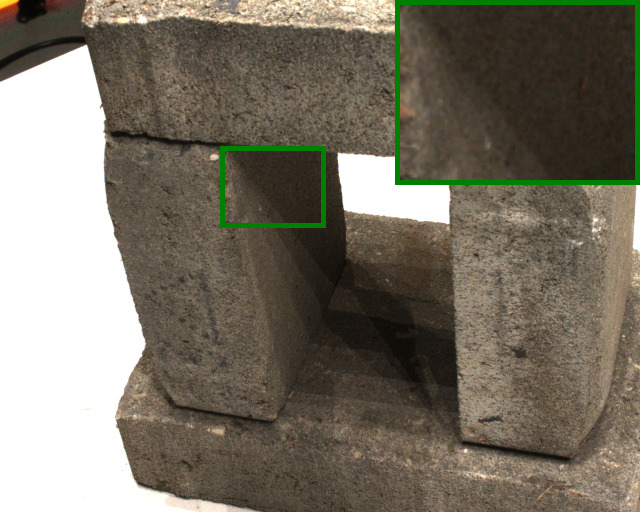}
    \end{subfigure}
    \begin{subfigure}[h]{\textwidth}
        \centering
        \includegraphics[width=0.19\linewidth]{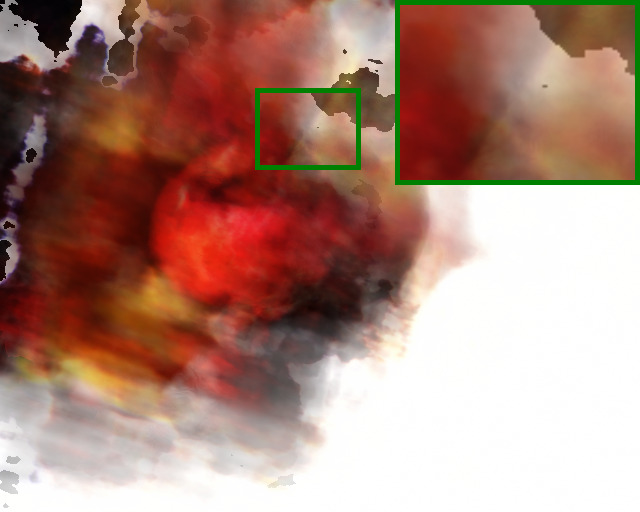}
        \includegraphics[width=0.19\linewidth]{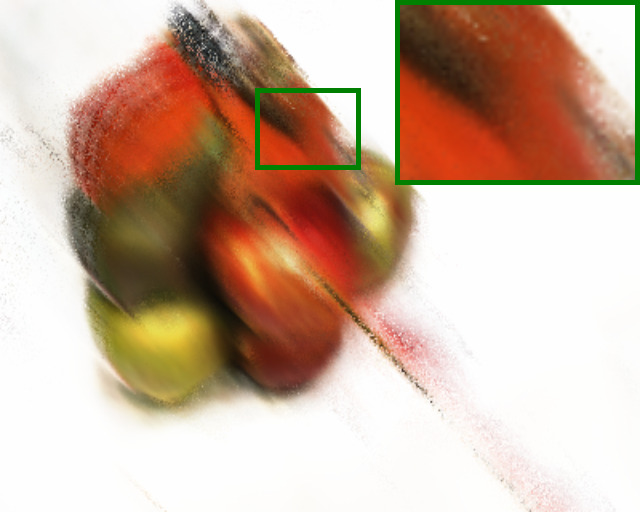}
        \includegraphics[width=0.19\linewidth]{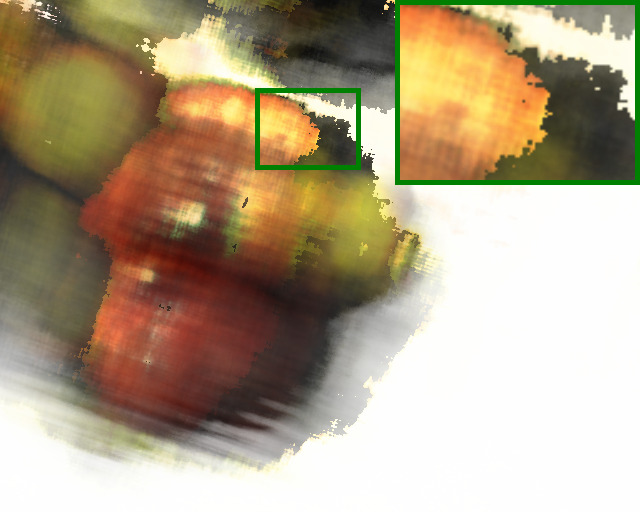}
        \includegraphics[width=0.19\linewidth]{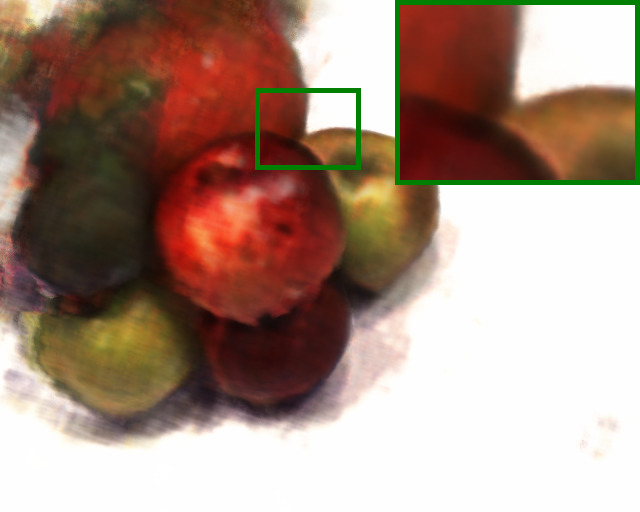}
        \includegraphics[width=0.19\linewidth]{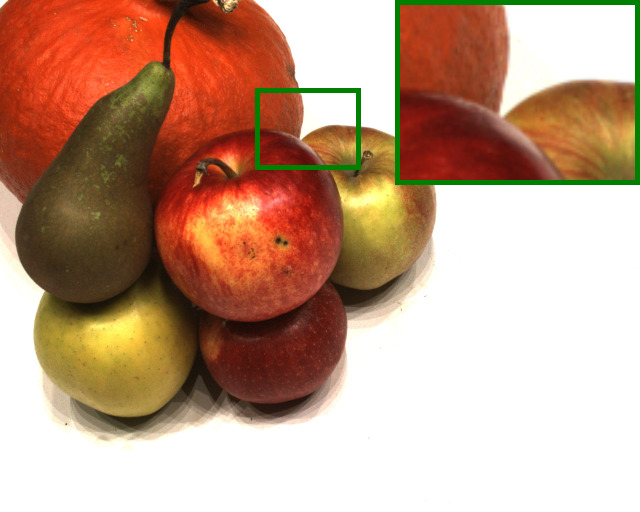}
    \end{subfigure}
   \begin{subfigure}[h]{\textwidth}
        \centering
        \includegraphics[width=0.19\linewidth]{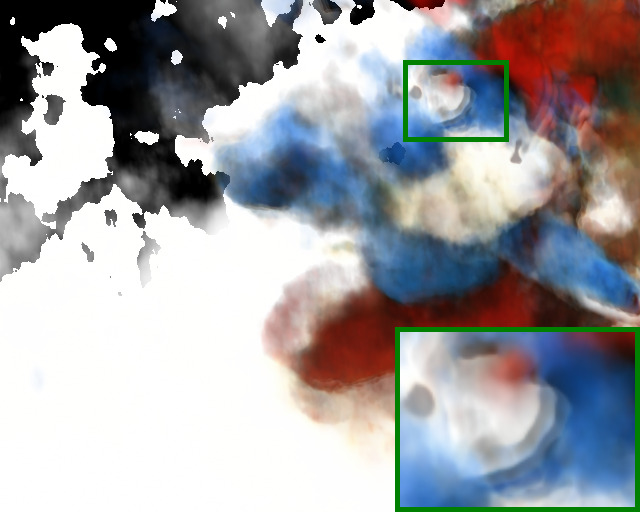}
        \includegraphics[width=0.19\linewidth]{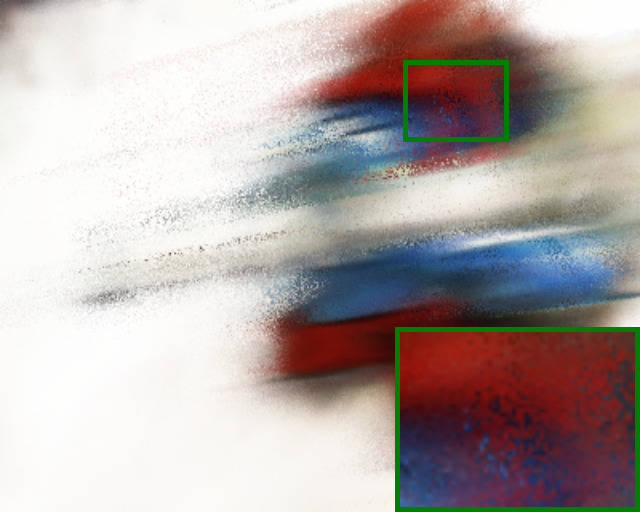}
        \includegraphics[width=0.19\linewidth]{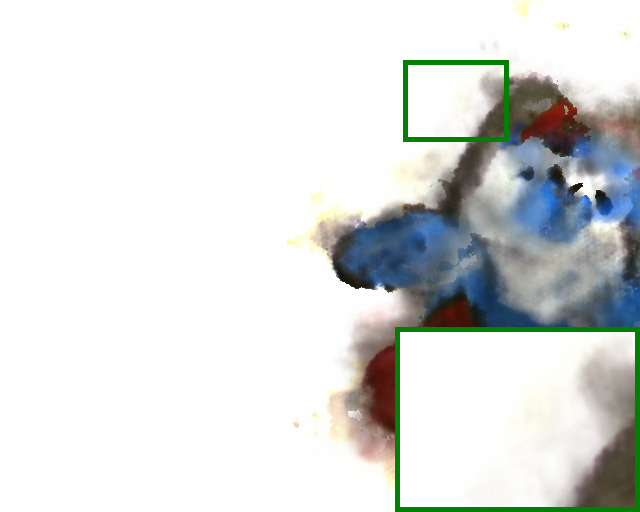}
        \includegraphics[width=0.19\linewidth]{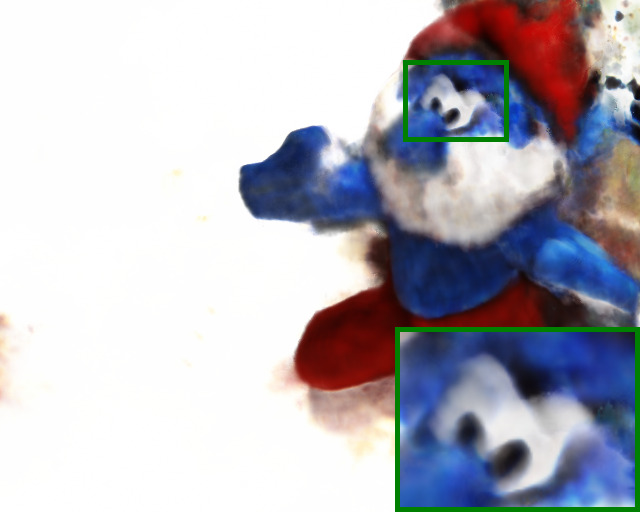}
        \includegraphics[width=0.19\linewidth]{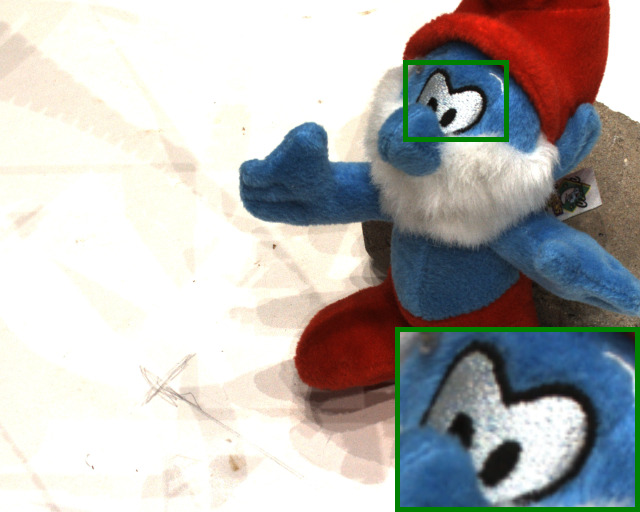}
    \end{subfigure}
    \begin{subfigure}[h]{\textwidth}
        \centering
        \includegraphics[width=0.19\linewidth]{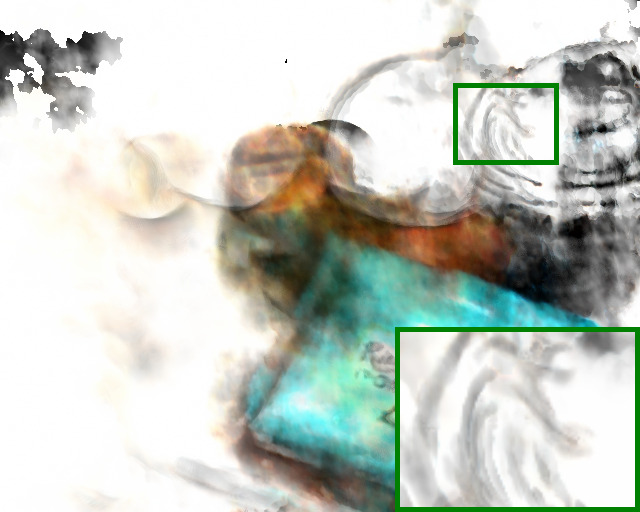}
        \includegraphics[width=0.19\linewidth]{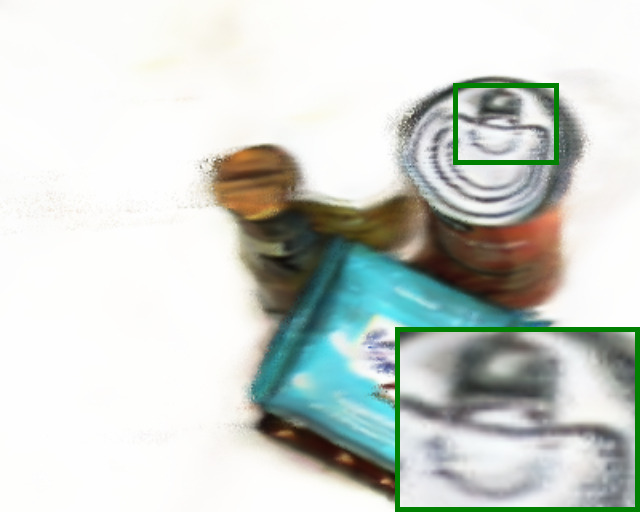}
        \includegraphics[width=0.19\linewidth]{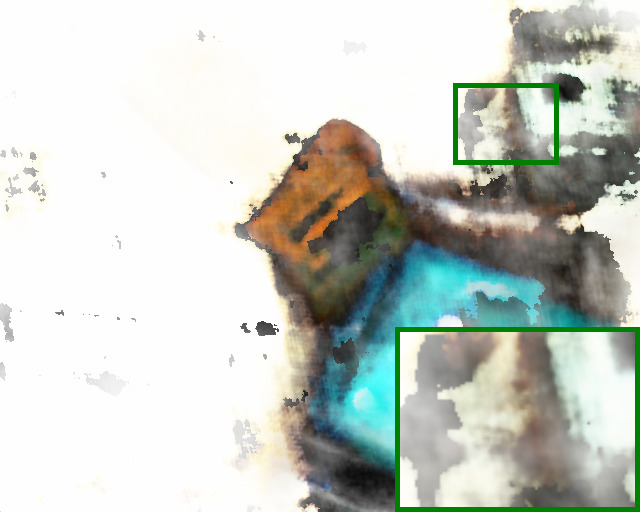}
        \includegraphics[width=0.19\linewidth]{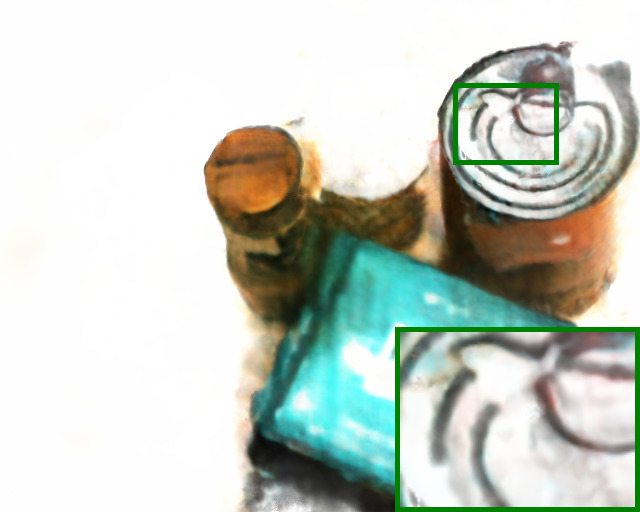}
        \includegraphics[width=0.19\linewidth]{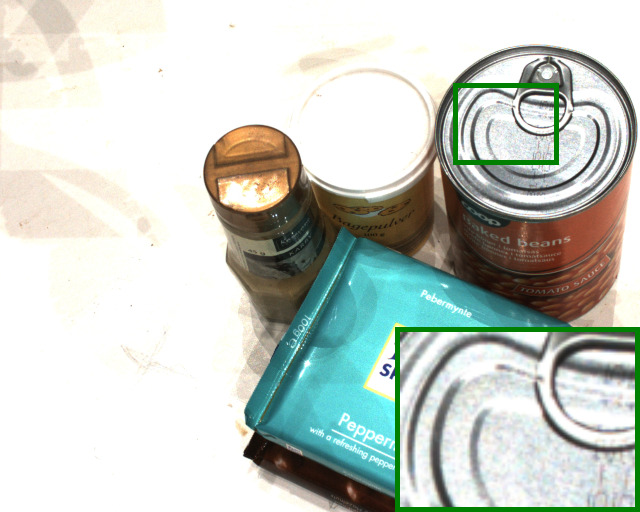}
    \end{subfigure}
    \begin{subfigure}[h]{\textwidth}
        \centering
        \includegraphics[width=0.19\linewidth]{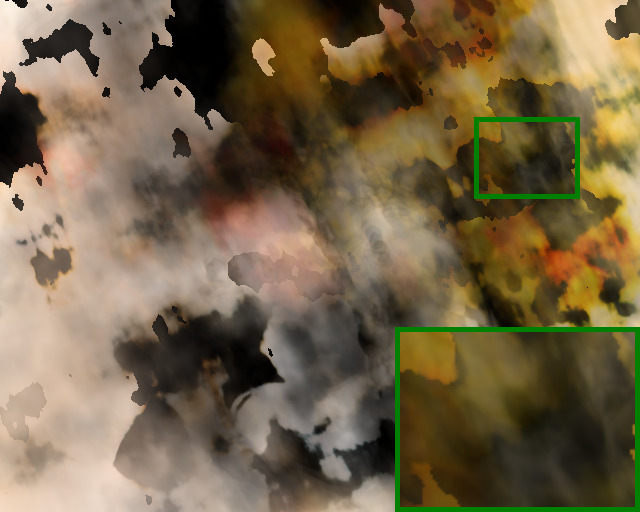}
        \includegraphics[width=0.19\linewidth]{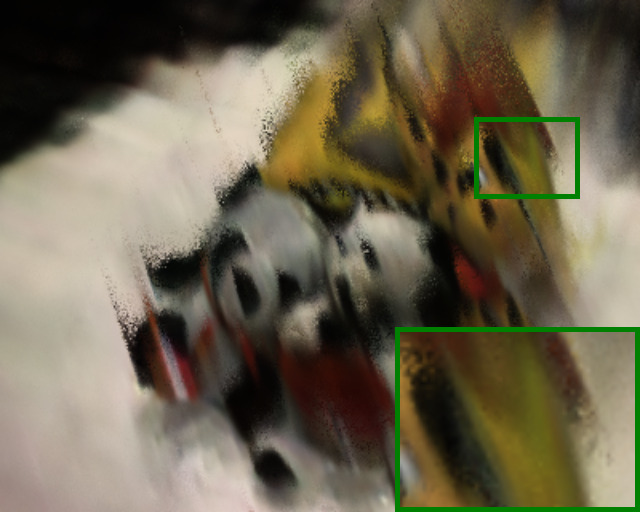}
        \includegraphics[width=0.19\linewidth]{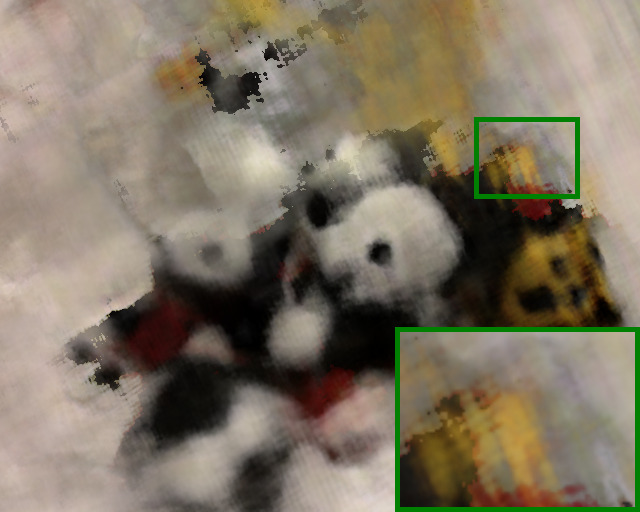}
        \includegraphics[width=0.19\linewidth]{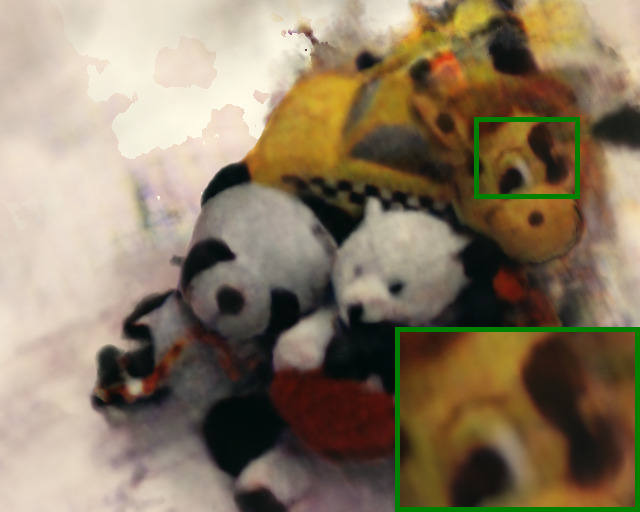}
        \includegraphics[width=0.19\linewidth]{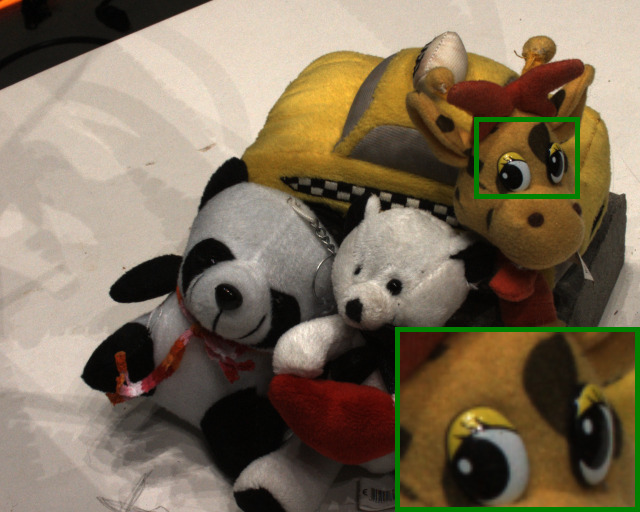}
    \end{subfigure}
\caption{Novel view synthesis results of different methods on DTU dataset.}\label{fig:dtu}
\end{figure}

\begin{table}[h]
\centering
\resizebox{\textwidth}{!}{%
\begin{tabular}{c|cccc|cccc|cccc}
 & \multicolumn{4}{c|}{PSNR$\uparrow$} & \multicolumn{4}{c|}{SSIM$\uparrow$} & \multicolumn{4}{c}{LPIPS$\downarrow$} \\
 & Lego & Hotdog & Flower & Room & Lego & Hotdog & Flower & Room & Lego & Hotdog & Flower & Room \\
 \hline
DS-NeRF & \cellcolor{orange!25}16.62 & \cellcolor{yellow!25}14.16 & \cellcolor{orange!25}16.92 & \cellcolor{orange!25}17.44 & \cellcolor{orange!25}0.77 & 0.67 & \cellcolor{orange!25}0.41 &\cellcolor{orange!25} 0.65 & \cellcolor{orange!25}0.1682 & 0.2956 & \cellcolor{orange!25}0.3900 & \cellcolor{orange!25}0.3986 \\
DietNeRF & \cellcolor{yellow!25}15.07 & \cellcolor{orange!25}16.28 & 13.35 & 15.77 & 0.72 & \cellcolor{yellow!25}0.69 & \cellcolor{yellow!25}0.20 & \cellcolor{yellow!25}0.49 & \cellcolor{yellow!25}0.2063 & \cellcolor{yellow!25}0.2633 & 0.7526 &\cellcolor{yellow!25} 0.7512 \\
PixelNeRF & 14.25 & 16.67 & \cellcolor{yellow!25}13.20 & \cellcolor{yellow!25}12.88 & \cellcolor{yellow!25}0.72 & \cellcolor{orange!25}0.71 & 0.19 & 0.41 & 0.2171 & \cellcolor{orange!25}0.2381 & \cellcolor{yellow!25}0.6378 & 0.7633 \\
\textbf{SinNeRF} & \cellcolor{red!25}20.97 & \cellcolor{red!25}19.78 & \cellcolor{red!25}17.20 & \cellcolor{red!25}18.85 & \cellcolor{red!25}0.82 & \cellcolor{red!25}0.77 & \cellcolor{red!25}0.41 & \cellcolor{red!25}0.67 & \cellcolor{red!25}0.0932 & \cellcolor{red!25}0.1700 & \cellcolor{red!25}0.3724 & \cellcolor{red!25}0.3796
\end{tabular}
}
\caption{Quantitative evaluation of our method against state-of-the-art methods on the NeRF synthetic dataset (Lego and Hotdog) and LLFF dataset (Flower and Room).}
\label{tab:synthetic}
\end{table}

\begin{table}[h]
\centering
 \begin{tabularx}{\linewidth}{>{\hsize=.9\hsize}X|>{\hsize=.9\hsize\centering\arraybackslash}X|>{\hsize=.9\hsize\centering\arraybackslash}X|>{\hsize=.9\hsize\centering\arraybackslash}X}
 & PSNR$\uparrow$ & SSIM$\uparrow$ & LPIPS$\downarrow$ \\ \hline
DS-NeRF & \cellcolor{yellow!25} 12.17 &	0.41 &	0.6493  \\
DietNeRF & \cellcolor{orange!25} 12.84 & \cellcolor{orange!25} 0.44 & \cellcolor{orange!25} 0.6469 \\
PixelNeRF &  12.06 & \cellcolor{yellow!25} 0.42 & \cellcolor{yellow!25}	0.6471  \\
\textbf{SinNeRF} & \cellcolor{red!25} 16.52	& \cellcolor{red!25} 0.56 &	\cellcolor{red!25} 0.5250
\end{tabularx}
\caption{Quantitative evaluation of our method against state-of-the-art methods on DTU dataset. We report average values across scenes.}
\label{tab:dtu}
\end{table}

\begin{figure}[h]
    \centering
    \begin{tabular}{P{0.18\textwidth}P{0.18\textwidth}P{0.18\textwidth}P{0.18\textwidth}P{0.18\textwidth}}
    \scriptsize w/o $\mathcal{L}_\text{adv}$ & \scriptsize w/o $\mathcal{L}_\text{cls}$ & \scriptsize w/o $\mathcal{L}_\text{geo}$ & \scriptsize Full Model & \scriptsize Target Image\\
    \end{tabular}
    \vskip\baselineskip
        \begin{subfigure}[b]{\textwidth}
        \centering
        \includegraphics[width=0.19\linewidth]{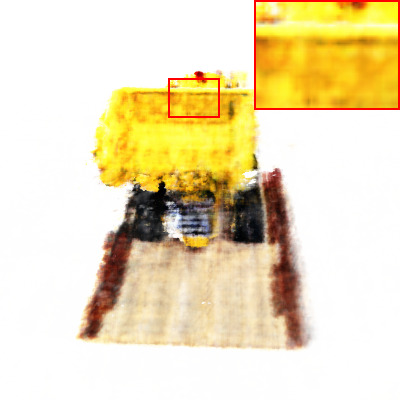}
        \includegraphics[width=0.19\linewidth]{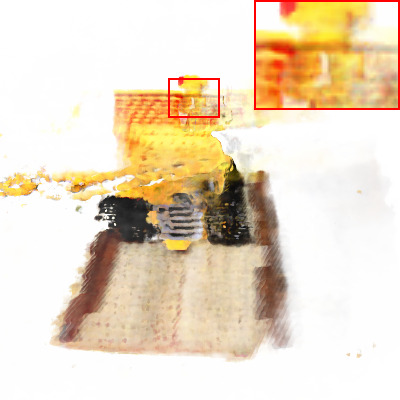}
        \includegraphics[width=0.19\linewidth]{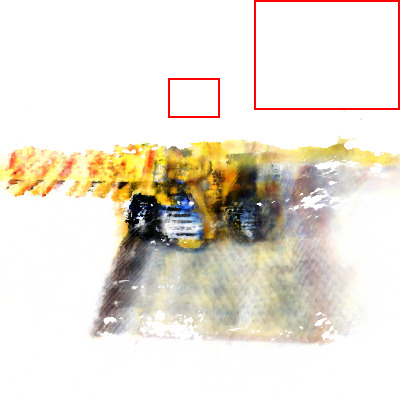}
        \includegraphics[width=0.19\linewidth]{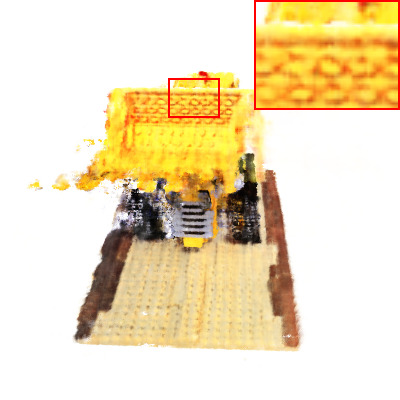}
        \includegraphics[width=0.19\linewidth]{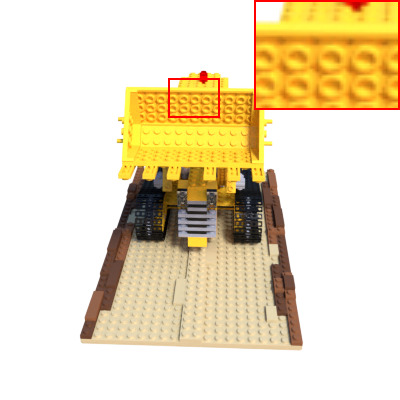}
    \end{subfigure}
    \caption{Novel view synthesis from different variants of our proposed model.}
    \label{fig:ablation}
\end{figure}

\subsection{Evaluation Protocol}

We perform experiments on NeRF synthetic dataset~\cite{nerf}, Local Light Field Fusion(LLFF) dataset~\cite{mildenhall2019local}, and DTU dataset~\cite{jensen2014large}.
NeRF synthetic dataset contains complex objects with $360^\circ$ view. LLFF provides complex forward-facing scenes. DTU consists of various objects placed on a table.
We report metrics including PSNR, structural similarity index (SSIM), and LPIPS perceptual metric~\cite{zhang2018perceptual}.
We compare our method with the state-of-the-art neural radiance field methods DietNeRF~\cite{jain2021putting}, PixelNeRF~\cite{yu2021pixelnerf}, and DS-NeRF~\cite{deng2021depth}.
We train DietNeRF and DS-NeRF for each scene since they are test-time optimization methods. As for PixelNeRF, we fine-tune the model on each scene before evaluation for a fair comparison.

\subsection{View synthesis on NeRF Synthetic Dataset}

For NeRF synthetic dataset, each scene is rendered via Blender. Both ground truth rendered images of 100 camera poses and the original blender files are provided. We randomly select a single view as the reference view and refer to its surrounding views as unseen views. Then we use blender to render the ground truth of the unseen views by rotating the world-to-camera matrix. Specifically, we generate 60 test set images by rotating the camera around the y-axis uniformly in $[-30^{\circ}, 30^{\circ}]$. The quantitative results are shown in Tab.~\ref{tab:synthetic}. Our method achieves the best results both in pixel-wise error and perceptual quality.

We show the novel view synthesis results in the first two rows of Fig.~\ref{fig:llff}. Each row corresponds to a fixed camera pose, and each column contains the results of a method. One can see that our method preserves the best geometry as well as perceptual quality. DS-NeRF's output contains a wrong geometry at the top of the lego. This is because DS-NeRF only utilizes supervision on the reference view and does not perform warping to other views. PixelNeRF's results contain ``ghost" hotdogs since they do not explicitly regularize the geometry. Optimizing on unseen views, DietNeRF produces appealing results, but unfortunately with flaws in the novel view's geometry (e.g., the objects are no longer in the center). The results are also blurry since their CLIP embeddings are obtained at a low resolution.

\subsection{View synthesis on LLFF Dataset}

For the local light field dataset, the images and the SfM results from colmap are provided. We randomly select a single view as the reference view and use its surrounding views as unseen views during training. For quantitative evaluation, we render the other views in the dataset whose ground truth images are available. We provide visual results in the last two rows of Fig.~\ref{fig:llff} and quantitative results in Tab.~\ref{tab:synthetic}. Our method generates the most visually-pleasing results, while other methods tend to render obscure estimations on novel views. DS-NeRF shows realistic geometry, but the rendered images are blurry. PixelNeRF and DietNeRF present good structures but wrong geometry due to their lack of local texture guidance and geometry pseudo label.

\begin{table}[tp]
\centering
 \begin{tabularx}{\linewidth}{>{\hsize=.9\hsize}X|>{\hsize=.9\hsize\centering\arraybackslash}X|>{\hsize=.9\hsize\centering\arraybackslash}X|>{\hsize=.9\hsize\centering\arraybackslash}X}
Methods & PSNR$\uparrow$ & SSIM$\uparrow$ & LPIPS$\downarrow$ \\ 
\hline
w/o $\mathcal{L}_\text{geo}$ & 16.11 \textcolor{amber}{(-4.86)} & 0.74 \textcolor{amber}{(-0.08)} & 0.1919 \textcolor{amber}{(+0.0987)} \\
w/o $\mathcal{L}_\text{cls}$ &  18.20 \textcolor{darkspringgreen}{(-2.77)} & 0.76 \textcolor{darkspringgreen}{(-0.06)} & 0.1348 \textcolor{red}{(+0.0146)}\\
w/o $\mathcal{L}_\text{adv}$ & 20.20 \textcolor{red}{(-0.77)} & 0.79 \textcolor{red}{(-0.03)} & 0.1306 \textcolor{darkspringgreen}{(+0.0294)} \\
Full Model & \textbf{20.97} & \textbf{0.82} & \textbf{0.0932} \\ 
\end{tabularx}
\caption{\textbf{Ablation study on variants of pseudo labels}.  ``w/o $\mathcal{L}_\text{adv}$" refers to the variant without the local texture guidance. ``w/o $\mathcal{L}_\text{cls}$" refers to the variant without global structure prior. ``w/o $\mathcal{L}_\text{geo}$" refers to removing the geometry pseudo labels and using depth supervision only on the reference view. Experiments are conducted on Lego scene.}
\label{tab:ablation}
\end{table}

\begin{table}[h]
\centering
 \begin{tabularx}{\linewidth}{>{\hsize=.9\hsize}X|>{\hsize=.9\hsize\centering\arraybackslash}X|>{\hsize=.9\hsize\centering\arraybackslash}X|>{\hsize=.9\hsize\centering\arraybackslash}X}
 & PSNR$\uparrow$ & SSIM$\uparrow$ & LPIPS$\downarrow$ \\ \hline
style & 15.49 \textcolor{amber}{(-5.48)} & 0.73 \textcolor{amber}{(-0.09)}& 0.2046 \textcolor{amber}{(+0.1114)}\\
self-similarity & 18.67 \textcolor{darkspringgreen}{(-2.30)}& 0.81 \textcolor{red}{(-0.01)} & 0.1075 \textcolor{darkspringgreen}{(+0.0143)}\\
content & 19.20 \textcolor{red}{(-1.76)}& 0.80 \textcolor{darkspringgreen}{(-0.02)}& 0.1138 \textcolor{red}{(+0.0206)}\\\relax
 \textbf{$[CLS]$ (ours)} & \textbf{20.97} & \textbf{0.82} & \textbf{0.0932}\\ 
\end{tabularx}
\caption{\textbf{Ablation study on different choices of the global structure prior.} Here  ``content loss" refers to calculating $L_1$ loss on the feature space of pretrained VGG-16 network~\cite{simonyan2014very}. ``style loss" refers to minimizing gram matrix from the output of pre-trained VGG-16 network~\cite{simonyan2014very}. ``self-similarity loss"~\cite{tumanyan2022splicing} refers to calculating the self-similarity of the keys in ViT's self-attention layer. The $[CLS]$ denotes our proposed one, where we adopt the $[CLS]$ token from pretrained DINO-ViT approaches. $\mathcal{L}_\text{cls}$. Experiments are conducted on Lego scene.}
\label{tab:cls_loss}
\end{table}

\subsection{View synthesis on DTU Dataset}

For each scene in DTU dataset, 49 images and their fixed camera poses are provided. We use camera 2 as the reference view because its images contain most parts of the scene. We use 10 nearby cameras from the dataset as unseen views during training. Since the ground truth of these nearby views are provided, we render these views for quantitative evaluation. We provide visual results in Fig.~\ref{fig:dtu} and quantitative results in Tab.~\ref{tab:dtu}. Our method demonstrates the most visually-pleasing results as well as the best quantitative performance. DS-NeRF generates realistic geometry, but the results contain severe artifacts. PixelNeRF and DietNeRF obtain a pleasing overall looking but suffer from wrong geometry.

\subsection{Ablation Study}

\paragraph{\textbf{Variants of pseudo labels}}. 
In this section, we study the effectiveness of each component of our proposed method.
We evaluate on the lego scene and provide the results in Fig.~\ref{fig:ablation} and Tab.~\ref{tab:ablation}.
Removing adversarial training leads to blurry artifacts. This is because the $\mathcal{L}_\text{cls}$ is only beneficial when the extracted patch has a receptive field large enough to cover the major structure of the image. The variant without global structure prior contains wrong structure in novel views, which is due to the missing guidance on the overall semantic structure. Although there are still geometry pseudo labels available, the projected depth information only provides partial guidance and leaves the occluded regions unconstrained. Finally, the variant without geometry pseudo labels suffers from wrong geometry. There is only depth supervision of the reference view, and the unseen views are not properly regularized.

\paragraph{\textbf{Different choices of the global structure prior}.}
We study different model choices for our global structure prior in this section. The global structure prior is designed to focus on the overall semantic consistency between the unseen views and the reference view regardless of the pixel misalignment. Following this direction, we evaluate different architectures including both the ConvNets and ViTs.
As shown in Tab.~\ref{tab:cls_loss}, we evaluate different kinds of the global structure prior by conducting experiments on the ``lego'' scene, including adopting the content, style, and self-similarity losses from a pre-trained VGG network between unseen views and reference view or minimizing the distance between the outputs of $[CLS]$ token from DINO-ViT~\cite{caron2021emerging} architecture. The quantitative results demonstrate that DINO-ViT shows a stronger global structure prior, suggesting that it is more robust to pixel misalignment.

\section{Conclusions}
We present SinNeRF, a framework to train a neural radiance field on a single view from a complex scene. SinNeRF is based on a semi-supervised framework, where geometry pseudo label and semantic pseudo label are synthesized to stabilize the training process. Comprehensive experiments are conducted on complex scene datasets, including NeRF synthetic dataset, Local Light Field Fusion (LLFF) dataset, and DTU dataset, where SinNeRF outperforms the current state-of-the-art NeRF frameworks. However, similar to most NeRF approaches, one limitation of SinNeRF is the training efficiency issue, which could be one of our future directions to explore further.

\pagebreak
\bibliographystyle{splncs04}
\bibliography{egbib}
\end{document}